\documentclass{article} 
\usepackage{iclr2027_conference,times}

\usepackage{amsmath,amsfonts,bm}

\def\eqref#1{equation~\ref{#1}}

\def\1{\bm{1}}

\DeclareMathAlphabet{\mathsfit}{\encodingdefault}{\sfdefault}{m}{sl}
\SetMathAlphabet{\mathsfit}{bold}{\encodingdefault}{\sfdefault}{bx}{n}

\usepackage{hyperref}
\usepackage{url}

\usepackage{xcolor}
\usepackage{colortbl}
\usepackage{comment}
\usepackage{booktabs}
 \usepackage{graphicx}

\title{Compiling Learning Problems into Adaptation Programs for Language Models}

\author{Rebecca Ramnauth \& Brian Scassellati \\
Department of Computer Science\\
Yale University\\
New Haven, CT 06511, USA \\
\texttt{rebecca.ramnauth@yale.edu}
}

\iclrfinalcopy 
\begin{document}

\maketitle

\begin{abstract}
Model adaptation is typically governed by a fixed recipe, even though different update programs can produce substantially different behavioral outcomes. We introduce adaptation compilation, which reframes where, how, and to what extent a model should adapt as a joint prediction and decision problem. Rather than searching over candidate programs anew for each learning episode, a compiler learns from prior adaptations to predict a vector-valued counterfactual response surface over candidate programs---their expected effects on acquisition, transfer, boundedness, and preservation---and selects a program before adaptation begins. Because this predicted geometry captures multiple behavioral consequences rather than a single winner or scalar score, it can be reused under different downstream priorities without retraining. Across five learning types, preferred programs vary meaningfully across episodes, and this variation is predictable from pre-adaptation information. On Llama-3.1-8B, compiler-selected programs approach exhaustive search while outperforming global and objective-specific defaults. Replication on Gemma-2-9B preserves program heterogeneity and selection headroom, but shows that exploiting this headroom requires accounting for uncertainty when departing from strong defaults. Together, these results show that adaptation search can be amortized across related learning problems, turning prior adaptation experience into a basis for deciding how future learning should occur.

\end{abstract}




\section{Introduction}
Current parameter-efficient methods begin from the assumption that before learning starts, one must decide where the model is allowed to change. Low-rank adaptation (LoRA) modules are commonly inserted into a fixed set of attention or feed-forward projections, often uniformly across depth. The rank, placement, regularization of these modules may be tuned, but the underlying adaptation architecture is typically specified independently of the learning problem itself. Different forms of learning (such as acquiring a new fact, applying a behavioral rule, planning a multistep procedure, or inferring a causal relationship) may therefore be implemented through the same update mechanism, even though they impose fundamentally different demands on the model.

This assumption has become increasingly difficult to justify. Different forms of learning are not equally well supported by updates to the same regions of a network \citep{ramnauth2026localized}. Some learning problems are acquired most effectively through highly localized updates, while others require broader or differently positioned changes. Moreover, the update location that maximizes immediate acquisition may not be the one that best promotes generalization---or, conversely, best prevents the learned behavior from extending beyond its intended scope. 

Recent methods partially relax fixed adaptation architecture in that they dynamically reallocate rank, prune unnecessary parameters, select modules using local sensitivity measures, or route inputs among previously learned adapters. However, these approaches tend to optimize capacity within a predefined adaptation scheme or respond to optimization signals during a single training run. They do not yet treat the adaptation architecture itself as something to be learned. We introduce this gap as \textbf{adaptation compilation}. That is, given the nature of an update, what parts of the model should change, what form should those changes take, and how should the resulting updates be constrained? 

A compiler receives a learning problem and produces an executable adaptation program. The program specifies where updates should occur, which modules should receive those updates, how much capacity each update should have, and how strongly unaffected behavior should be preserved. Adaptation is then performed only within the compiled program. Rather than adapting a model through a static configuration and evaluating its consequences afterward, the system predicts how a learning episode is expected to respond to alternative adaptation configurations and selects the configuration expected to produce the intended behavioral profile. 

This matters for system reliability because such a compiler is not only efficient---avoiding exhaustive adaptation sweeps or costly diagnostics for every update---but also distinguishes whether an update was acquired, generalized appropriately, remained bounded, and preserved unrelated behavior. A lightweight system could therefore characterize an episode, predict the consequences of candidate programs, and reserve more expensive probing or explicit search for ambiguous cases. We develop this idea through six experiments: first establishing whether meaningful selection headroom exists; then testing whether adaptation outcomes can be predicted from pre-adaptation information and used to select better programs; identifying which episode–model signals matter; and finally evaluating generalization to unseen learning types and replication on a second model backbone.\footnote{We provide the code, dataset, experimental outputs, and supporting artifacts for reproducibility in the \href{https://github.com/rramnauth2220/adaptcompile-experiments}{supplementary GitHub repository}, along with the \href{https://pypi.org/project/adaptcompile/}{\texttt{adaptcompile}} Python package, which exposes adaptation compilation as a reusable interface for future applications and extensions.}

\section{Background and Related Work}

Adapting a pretrained model requires two core decisions: (1) how its parameters should change, and (2) which degrees of freedom should be available for adaptation. Our work connects prior research on parameter-efficient design, automated allocation, meta-learning, and task representation by treating adaptation structure as something that can be predicted from a new learning episode.


Parameter-efficient fine-tuning (PEFT) restricts adaptation to a small set of trainable degrees of freedom, including adapters, prompts, selected pretrained parameters, or low-rank updates \citep{houlsby2019parameter,li2021prefix,lester2021power,zaken2022bitfit,hu2022lora}. These methods therefore require an adaptation configuration specifying where and how learning may occur. Such configurations are commonly fixed before optimization, even though effective update locations can vary across learning objectives and need not optimize acquisition, transfer, and boundedness simultaneously \citep{ramnauth2026localized}. This motivates treating adaptation structure as part of the learning problem rather than solely as a manually specified hyperparameter.




Several approaches already make PEFT structure adaptive. Dynamic rank allocation, architecture search, and model-derived diagnostics can redistribute capacity or identify promising layers and modules \citep{zhang2023adalora,valipour2023dylora,mao2024dora,hu2022sparse,lawton2023neural,zhou2024autopeft,xu2026understanding,saket2026aletheia,zhang2026rethinking}. These methods establish that adaptation configuration is consequential, but generally optimize structure within the target training problem, search anew for each task, or translate a predefined diagnostic into an allocation rule. Our setting instead asks whether outcomes from \emph{previous} adaptations can be reused to predict the consequences of alternative configurations for a new episode before adaptation begins.




This perspective connects adaptation compilation to meta-learning and per-instance algorithm selection. Meta-learning learns initializations, optimization rules, or adaptable parameter sets across tasks \citep{finn2017model,li2017meta,andrychowicz2016learning,ravi2017optimization,von2021learning}, while Task2Vec and related approaches represent tasks through their interaction with a model rather than nominal task identity \citep{achille2019task2vec,vu2020exploring,wang2021grad2task}. Classical algorithm selection similarly uses features of a new problem instance to predict which candidate procedure will perform best \citep{rice1976algorithm}. To our knowledge, these ideas have not been combined to predict the multidimensional behavioral consequences of each candidate adaptation program so that program selection can depend on the behavioral tradeoff of interest rather than directly predicting a single winner.

\section{A Framework for Adaptation Compilation} \label{sec:framework}

We formulate this missing capability as \textbf{adaptation compilation}, which predicts the consequences of alternative update programs for a new learning episode and selects among them before adaptation begins. \citet{ramnauth2026localized} introduced \textit{adaptation geometry} and characterized it through empirical search over adaptation configurations; here, we amortize that search by learning to predict geometry from prior adaptation episodes.

Let $M_\theta$ denote the pretrained model and $D$ the adaptation examples defining a desired model change. Let $V(M)$ be the set of modules that could be adapted. An adaptation configuration
\[
g=\{(v,r_v):v\in V_g\subseteq V(M)\}
\]
specifies the adaptable modules $v$ and their update capacities $r_v$; all other modules outside $V_g$ are frozen. We restrict candidate programs to those satisfying a cost budget $C(g)\leq C_{\max}$.

Executing program $g$ on episode $D$ produces behavioral outcomes
\[
Y_{D,M}(g;\xi)=[A,T,B,P],
\]
where $\xi$ captures optimization randomness and $A$, $T$, $B$, and $P$ measure acquisition, transfer, boundedness, and preservation.\footnote{
Following \citet{ramnauth2026localized}, \textit{acquisition} measures whether the intended change is learned, \textit{transfer} whether it extends to appropriate held-out contexts, and \textit{boundedness} whether it is withheld where it should not apply. We add \textit{preservation} to measure whether unrelated pre-existing behavior remains unchanged. Cost is treated separately because it describes the adaptation procedure rather than its behavioral consequences.
} The expected outcomes across adaptation randomness define the episode's \textit{adaptation geometry},
\[
\Gamma_{D,M}(g)=\mathbb{E}_{\xi}[Y_{D,M}(g;\xi)].
\]
Thus, $\Gamma_{D,M}$ describes how the same learning episode is expected to behave under alternative adaptation programs.

Before adaptation, we characterize the episode and its interaction with the frozen model by
\[
s_{D,M}=[e_D,b_{D,M},q_{D,M}],
\]
where $e_D$ represents the adaptation examples, $b_{D,M}$ summarizes the frozen model's behavior on them, and $q_{D,M}$ contains module-level diagnostics. A learned predictor
\[
F_\psi(s_{D,M},g)=\widehat{\Gamma}_{D,M}(g)
\]
estimates the behavioral geometry of each candidate program from records collected on previous adaptation episodes.

Given a utility specification $\tau$ describing the desired tradeoff among behavioral outcomes, the compiler selects
\[
g^*=\arg\max_{g\in\mathcal G(M),\,C(g)\leq C_{\max}}
U_\tau\!\left(\widehat{\Gamma}_{D,M}(g)\right).
\]
The selected $g^*$ is the \textit{adaptation program}, the executable specification of where and how adaptation will occur. Compilation happens before adaptation, after which the selected program is trained through ordinary parameter-efficient optimization.

\section{Experimental Setup} \label{sec:methods}
We evaluate adaptation compilation through five questions that follow the logic of the compiler itself: whether alternative programs create meaningful selection headroom (RQ1), whether their behavioral consequences can be predicted before adaptation (RQ2), whether those predictions support better program selection (RQ3), which pre-adaptation signals enable that prediction (RQ4), and whether the learned relationships generalize beyond represented learning families (RQ5). The following setup is shared across the five corresponding experiments (Sec.~\ref{sec:exp1}--\ref{sec:exp5}).

\textbf{Learning Episodes.} We build on the benchmark introduced by \citet{ramnauth2026localized}, using the same latent-specification framework, example-generation procedure, and five learning objectives: lexical binding, factual association, behavioral policy learning, causal mapping, and procedural reasoning. Each latent specification defines an episode with an adaptation set and held-out evaluations for the behavioral outcomes defined in Sec.~\ref{sec:framework}; preservation is evaluated on examples unrelated to the target adaptation. Episodes are partitioned at the latent-specification level into 400 meta-training, 100 validation, and 100 test episodes, balanced across objectives. Objective identity is not provided to the geometry predictor.



\textbf{Model backbone.}
All experiments use Llama-3.1-8B-Instruct \citep{grattafiori2024llama}, the primary backbone used by \citet{ramnauth2026localized}. We retain it for continuity with the benchmark and computational tractability under repeated LoRA interventions.


\textbf{Configuration space.}
We instantiate $\mathcal{G}(M)$ using four approximately budget-matched LoRA programs: early-, middle-, and late-depth rank-$16$ adaptation, and full-stack rank-$4$ adaptation. Layer position is represented by normalized model depth.


\textbf{Utility.}
We use $U_\tau=\mathbf{w}_\tau^\top[A,T,B,P]/\|\mathbf{w}_\tau\|_1$, with
$\mathbf{w}_{\mathrm{bal}}=(1,1,1,1)$ and transfer-, boundedness-, and preservation-heavy variants $(1,2,1,1)$, $(1,1,2,1)$, and $(1,1,1,2)$, respectively. Because candidate programs are approximately budget matched, cost does not enter the experimental utility.

\textbf{Optimization schedule calibration.} 
We calibrate the optimization schedule on meta-training episodes using full-stack rank-$16$ LoRA as a high-capacity reference, then fix the selected schedule across candidate programs and evaluation splits. Full details are provided in Appendix~\ref{app:calibration}.

\textbf{Adaptation records.}
Each candidate program is independently executed for every episode and evaluated on acquisition, transfer, boundedness, and preservation. Meta-training and validation episodes use one adaptation seed; test episodes use three, whose outcomes are averaged to estimate expected geometry. Test outcomes are never available before compiler selection.

\textbf{Episode--model features.}
Before adaptation, we compute (1) pooled frozen hidden-state representations of the adaptation examples, (2) frozen-model target-token loss statistics, and (3) module-level probes of sensitivity, gradient magnitude, cross-example gradient agreement, and activation magnitude across module types and normalized depth. All features are computed before adaptation, and none requires training a candidate program.


\textbf{Geometry predictor.}
We fit a predictor $F_\psi(s_{D,M},g)$ to estimate acquisition, transfer, boundedness, and preservation for each episode--program pair. The full episode--model representation was pre-specified as the primary representation; within it, we compare ridge regression and random-forest regression and select the model and hyperparameters by validation mean absolute error (MAE), averaged across the four behavioral outcomes. Representation ablations are evaluated separately and are not candidates for primary-model selection. The test set is held out until final evaluation.

\section{Experiment 1: Is Adaptation Compilation Necessary?} \label{sec:exp1}

Before attempting to predict adaptation geometry, we first ask whether there is a meaningful program-selection problem to solve (RQ1). We consider two prerequisites for compilation. First, \textit{configuration sensitivity} asks whether alternative adaptation programs produce meaningfully different outcomes for the same episode. Second, \textit{selection headroom} asks whether the preferred program varies sufficiently across episodes to justify episode-conditioned selection. We compare the episode oracle with a \textit{global-fixed} policy
learned from meta-training episodes and an \textit{objective-fixed} policy that selects one program per learning objective. The remaining gap between the objective-fixed policy and the oracle therefore measures headroom beyond objective identity alone.

\begin{figure}[t]
    \centering
    \includegraphics[width=0.9\linewidth]{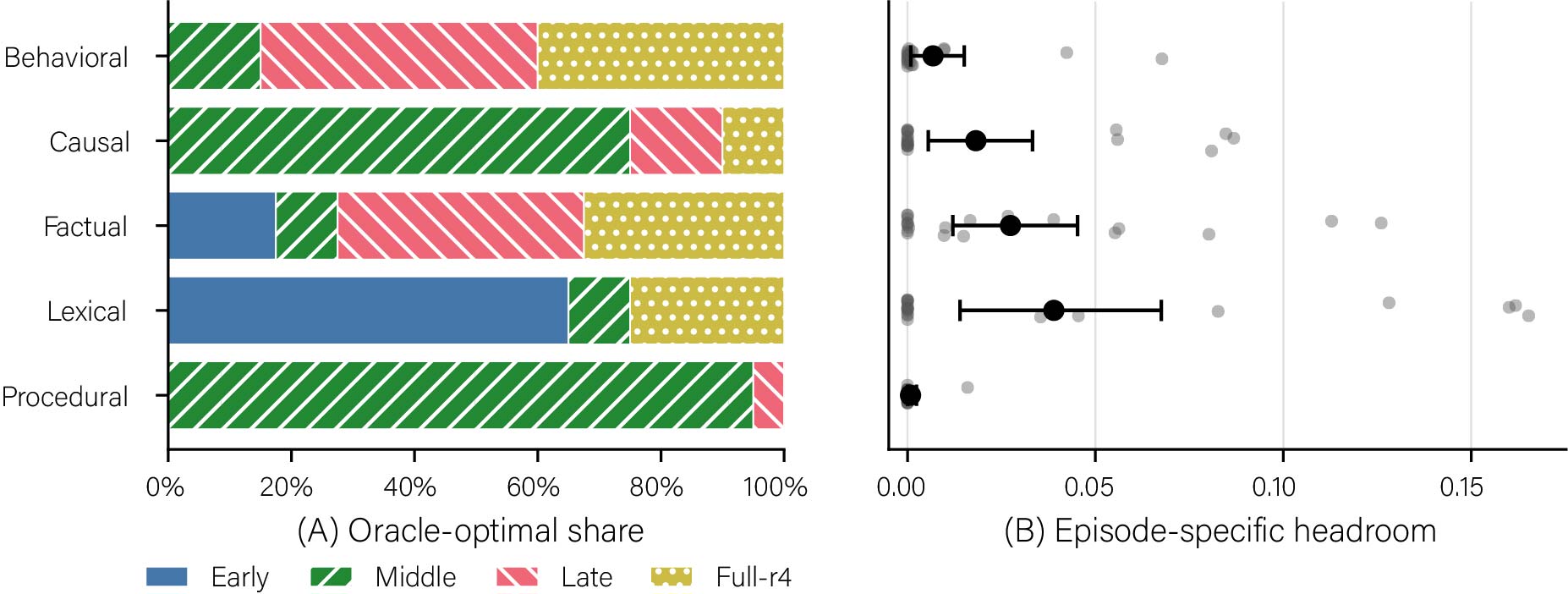}
    \caption{
    \textbf{Selection headroom in adaptation geometry.}
    \textbf{(A)} Fraction of held-out episodes for which each adaptation program is oracle-optimal, shown by learning objective.
    \textbf{(B)} Episode-specific headroom beyond the objective-fixed policy, measured as oracle utility minus objective-fixed utility. Gray points denote episodes; black points and error bars show the mean and 95\% CI. Headroom is largest for lexical binding and factual association and nearly absent for procedural reasoning.
    }
    \label{fig:exp1}
\end{figure}

On 100 held-out episodes, the global-fixed policy achieves mean utility $0.569$, objective-fixed selection improves this to $0.600$, and the episode oracle reaches $0.618$ (Fig.~\ref{fig:exp1}a).
Objective identity therefore explains substantial structure, but does not eliminate the selection problem. Specifically, objective-fixed selection remains oracle-optimal on only $64\%$ of episodes, compared with $41\%$ for the global policy. Moreover, $99\%$ of episodes have a unique utility-maximizing program,
and all four candidate programs are oracle-optimal for substantial subsets of episodes.

The remaining headroom is heterogeneous across learning objectives. It is largest for lexical binding and factual association, smaller for causal and behavioral learning, and nearly absent for procedural reasoning (Fig.~\ref{fig:exp1}b). Thus, some learning regimes admit strong structural defaults, whereas others retain meaningful episode-level variation in which program best balances acquisition, transfer, boundedness, and preservation. Additional analyses are provided in Appendix~\ref{app:exp1}.


\section{Experiment 2: Can Adaptation Geometry Be Predicted?} \label{sec:exp2}

Having established meaningful selection headroom, we next ask whether adaptation geometry can be predicted before adaptation (RQ2). For each held-out episode--program pair, the predictor estimates acquisition, transfer, boundedness, and preservation from pre-adaptation information. The full representation uses a random forest, selected over ridge regression by validation MAE ($0.044$ vs.~$0.099$). We compare against a configuration-mean baseline and, diagnostically, an objective-conditioned mean predictor given the true learning objective.

\begin{figure}[t]
    \centering
    \includegraphics[width=\linewidth]{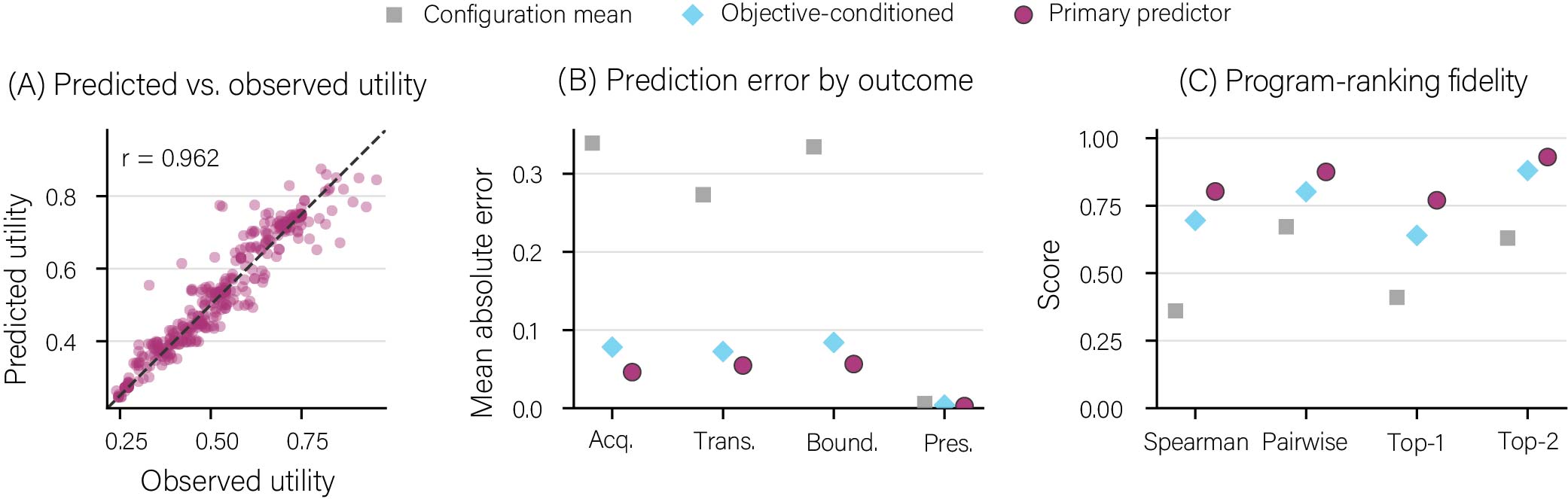}
    \caption{
    \textbf{Predicting adaptation geometry from pre-adaptation episode--model signals.}
    \textbf{(A)} Predicted and observed balanced utility across held-out episode--program pairs.
    \textbf{(B)} Mean absolute error for acquisition, transfer, boundedness, and preservation under the primary predictor and non-episode-conditioned baselines.
    \textbf{(C)} Within-episode program-ranking fidelity measured by Spearman correlation, pairwise ranking accuracy, and tie-aware top-1 and top-2 oracle recovery.
    }
    \label{fig:exp2}
\end{figure}

On held-out episodes, the primary predictor achieves geometry MAE of $0.040$, compared with $0.238$ for the configuration-mean baseline and $0.060$ for the objective-conditioned diagnostic (Fig.~\ref{fig:exp2}). Predicted and observed program utilities are also strongly aligned ($r=0.962$). More importantly for compilation, the predictor preserves within-episode program ordering, reaching Spearman correlation $0.803$, pairwise ranking accuracy $87.5\%$, and top-1/top-2 oracle recovery of $77\%$/$93\%$. These results show that pre-adaptation episode--model signals contain information beyond objective identity about how alternative programs will behave, including the relative ordering needed for selection. Additional per-objective analyses are provided in Appendix~\ref{app:exp2}.

\section{Experiment 3: Does Prediction Yield Better Programs?} \label{sec:exp3}

\begin{figure}[t]
    \centering
    \includegraphics[width=0.9\linewidth]{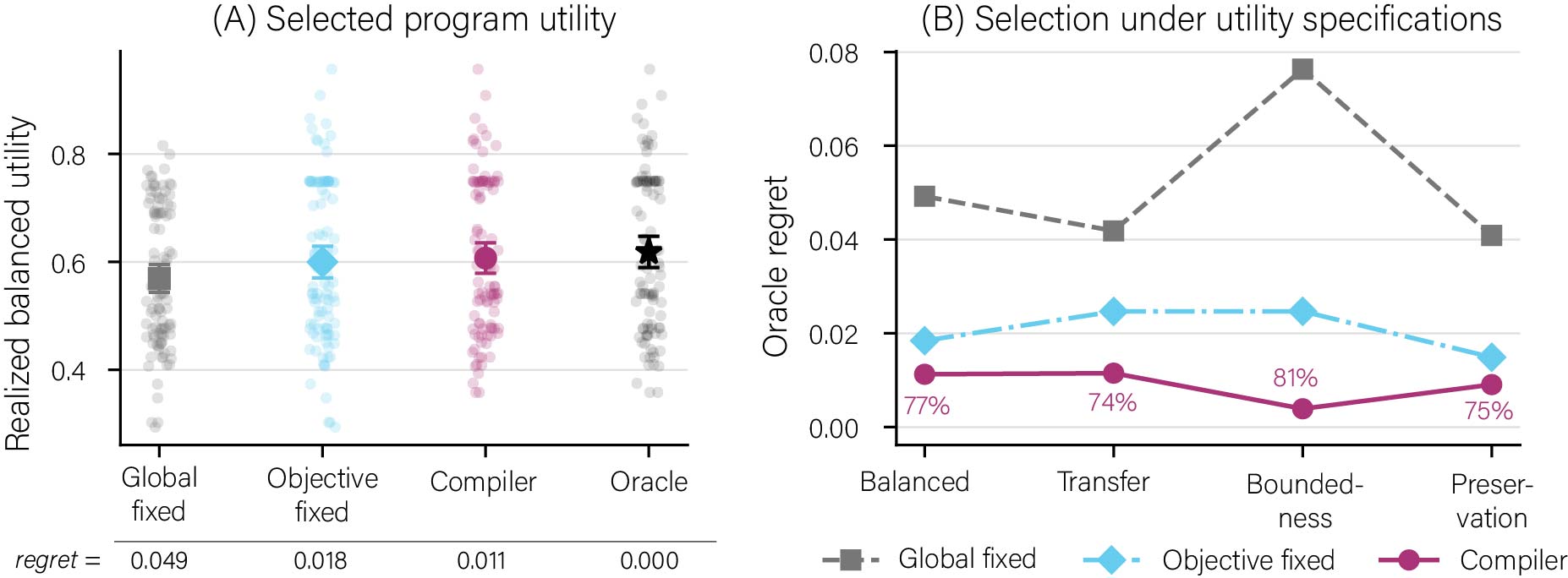}
    \caption{
    \textbf{Program selection from predicted adaptation geometry.}
    \textbf{(A)} Realized balanced utility for global-fixed, objective-fixed, compiler, and oracle selection; values below the axis show mean oracle regret.
    \textbf{(B)} Oracle regret under alternative utility specifications, using the same predicted geometry without retraining. Percentages show tie-aware top-1 oracle recovery.
    }
    \label{fig:exp3}
\end{figure}

We next ask whether predicted geometry supports better program selection (RQ3). Under balanced utility, the objective-fixed policy already captures much of the available selection headroom (Fig.~\ref{fig:exp3}a), achieving $0.600$ mean utility versus $0.618$ for the exhaustive oracle. The compiler reaches $0.607$, reducing oracle regret from $0.018$ to $0.011$ and recovering roughly $39\%$ of the remaining gap. Across 100 held-out episodes, compilation improves over objective-fixed selection on 22 episodes, degrades performance on 7, and yields identical realized utility on 71 (mean paired gain \(=0.0072\), 95\% bootstrap CI \([0.0011,0.0136]\)). Compiler regret remains below 0.006 for four of five learning objectives, with lexical binding the clearest remaining failure mode (Appendix~\ref{app:exp3}).


Because the predictor estimates behavioral outcomes rather than a single winner, the same predicted geometry can be recompiled under different priorities without retraining. The compiler remains closer to the oracle than either fixed policy under transfer-, boundedness-, and preservation-heavy utilities (Fig.~\ref{fig:exp3}b), showing that predicted geometry can support different adaptation objectives after prediction. Additional per-objective and utility-specific results are provided in Appendix~\ref{app:exp3}.

\section{Experiment 4: Which Pre-Adaptation Signals are Needed?} \label{sec:exp4}

We next ask how much pre-adaptation information is needed to predict geometry and support program selection (RQ4). Our primary representation combines adaptation-example representations, frozen-model behavioral statistics, and module-level probes. Surprisingly, the episode-only representation performs comparably to the full representation, achieving geometry MAE of $0.0434$ versus $0.0435$, with within-episode Spearman correlation of $0.848$ and top-1 oracle recovery of $83\%$.

Module probes and frozen-model behavioral statistics are weaker independently, with geometry MAE of $0.053$ and $0.054$, respectively. These results suggest that, in the present setting, most of the signal needed for compilation is already available in frozen representations of the learning episode. Explicit gradient and behavioral diagnostics therefore provide limited additional benefit, indicating that useful compilation may require less pre-adaptation computation than expected (Fig.~\ref{fig:compilation_scope}a). Full prediction and selection results are provided in Appendix~\ref{app:exp4}.

\section{Experiment 5: Does Adaptation Compilation Generalize?} \label{sec:exp5}

Experiments 2--4 evaluate unseen episodes drawn from learning families represented during meta-training. We finally ask whether the learned relationship between episode representations and adaptation geometry transfers to an entirely unseen learning family (RQ5). In leave-one-family-out evaluation, one objective is excluded entirely from both training and validation, and the resulting predictor is evaluated on test episodes from that unseen family.

\begin{figure}[t]
    \centering
    \includegraphics[width=\linewidth]{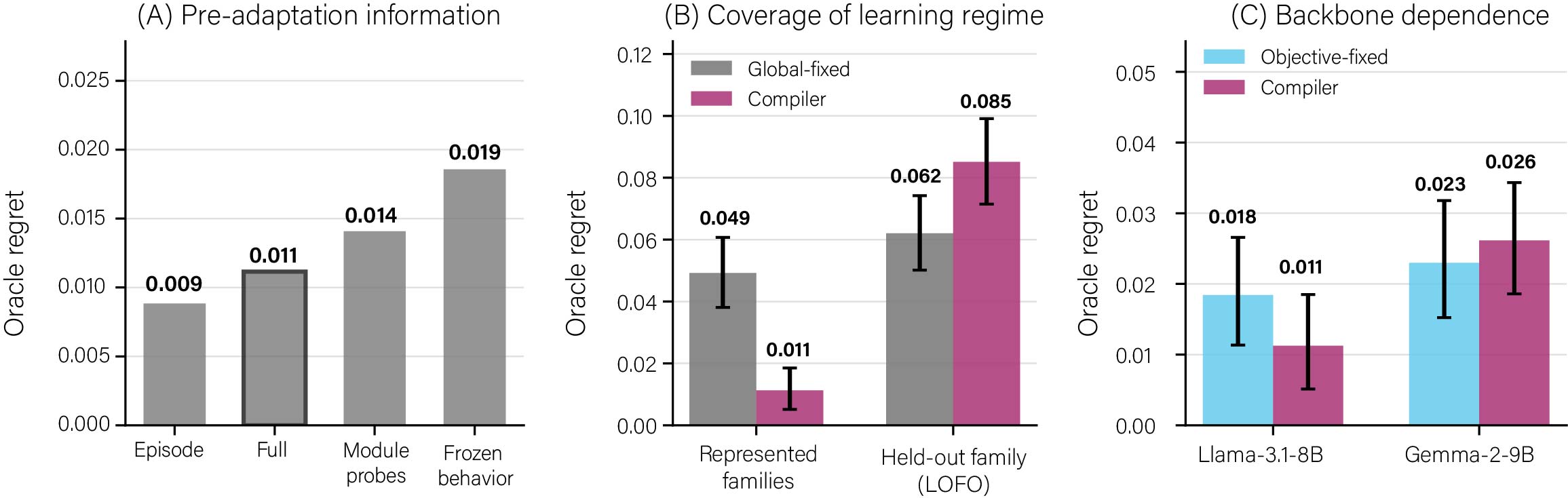}
    \caption{\textbf{Scope and boundary conditions of adaptation compilation.}
\textbf{(A)} Episode representations alone support selection comparable to the full representation, while module probes and frozen-behavior features are weaker in isolation; the outlined bar marks the primary full representation.
\textbf{(B)} Compilation reduces oracle regret when learning families are represented during meta-training, but this advantage reverses when an entire family is withheld (LOFO).
\textbf{(C)} Compilation improves over objective-fixed defaults on Llama, whereas on Gemma direct selection from predicted geometry does not improve over the strong objective-fixed policy despite remaining episode-level headroom. Error bars show 95\% episode-bootstrap CIs where applicable. Lower oracle regret is better.}
\label{fig:compilation_scope}
\end{figure}

Zero-shot family transfer is substantially more difficult than prediction for unseen episodes from represented objectives. Mean geometry MAE rises to $0.339$, within-episode Spearman correlation falls to $0.058$, and top-1 oracle recovery drops to $14\%$. Correspondingly, compiler utility falls to $0.533$, below the global-fixed policy at $0.556$ (Fig.~\ref{fig:compilation_scope}b). Performance is heterogeneous across families, but the overall result is that strong generalization to unseen episodes within represented learning regimes does not imply zero-shot transfer to entirely unseen forms of learning. Additional per-family results are provided in Appendix~\ref{app:exp5}.


These results establish an important boundary on the present form of adaptation compilation. The predictor generalizes well to new episodes within learning families represented during meta-training, but does not reliably extrapolate adaptation geometry to qualitatively unseen learning objectives. The learned mapping therefore captures structure that is more general than individual latent specifications, but remains dependent on coverage of the underlying learning regime.

\section{Experiment 6: Does the Case for Compilation Reproduce Across Backbones?} \label{sec:exp6}

Experiments 1--5 establish adaptation compilation on Llama-3.1-8B-Instruct, but adaptation geometry is inherently model-dependent. \citet{ramnauth2026localized} finds broadly consistent localization signatures across models, alongside substantial model-specific variation and particular sensitivity in Gemma to Llama-calibrated adaptation budgets. We therefore repeat the compiler pipeline on Gemma-2-9B-IT, independently recalibrating the optimization schedule while holding the program library and evaluation protocol fixed. The predictor is trained only on Gemma adaptation records, testing replication of the compilation principle rather than zero-shot transfer from Llama.

On Gemma, the best fixed program is full-stack. Objective-specific defaults select full-stack for behavioral, factual, and procedural learning, middle for causal mapping, and late for lexical binding; only the causal default matches Llama. Yet no program dominates held-out episodes: across 100 test episodes, oracle selections are 40\% full, 31\% middle, 23\% late, and 6\% early. The best fixed program achieves .436 mean utility, versus .461 for the episode-wise oracle; objective-specific defaults reach .438 but retain .023 mean regret. Thus, adaptation geometry changes across backbones, but substantial episode-level selection headroom remains even after conditioning on learning type.

Recovering this headroom is more difficult for Gemma. The learned compiler achieves .435 mean utility versus .461 for the oracle, with .026 regret, 42\% oracle recovery, and 73\% top-2 recovery. Its aggregate performance is comparable to, but does not exceed, the global-fixed (.436) or objective-fixed (.438) baselines (Fig.~\ref{fig:compilation_scope}c). The shortfall is concentrated, not uniform; the compiler largely recovers strong defaults for behavioral, causal, and factual episodes, with most loss arising in procedural reasoning and a smaller deficit in lexical binding. In these cases, small predicted differences between candidate programs can trigger a switch even when the realized advantage is uncertain.

Gemma sharpens rather than weakens the case for compilation. Program heterogeneity and episode-level oracle headroom persist across backbones, but exploiting that headroom requires knowing when a predicted advantage over a strong default is reliable. On Llama, direct selection from predicted geometry improves over fixed policies; on Gemma, the same rule sometimes acts on small, noisy margins. This suggests an additional design requirement for adaptation compilers. Program selection should account not only for predicted geometry, but also for uncertainty in the decision to depart from a robust default.

\section{Discussion}
Adaptation compilation begins from a methodological mismatch: structural choices---where to update, which modules to expose, and how much capacity to allocate---are typically fixed before knowing how a given learning episode will respond. Prior work shows that these choices induce distinct behavioral profiles across learning objectives  \citep{ramnauth2026localized}, but identifying those profiles ordinarily requires running the very adaptations one hopes to choose among. We ask whether this search can instead be amortized across prior adaptation experience.

Our results show that it can, but also clarify the conditions under which doing so is useful. On Llama, approximately budget-matched programs produce meaningfully different outcomes across episodes, those differences are predictable before adaptation, and selecting from predicted geometry substantially closes the gap to exhaustive search while outperforming global and objective-specific defaults. The same predictions can also be reused under different behavioral priorities because the compiler estimates acquisition, transfer, boundedness, and preservation rather than directly predicting a single winning program. Experiment~6 shows that the case for compilation reproduces on Gemma: adaptation geometry changes across backbones, yet substantial episode-level selection headroom remains. At the same time, direct selection from predicted geometry is less reliable on Gemma, revealing that headroom alone is insufficient; a useful compiler must also know when a predicted advantage over a strong default is trustworthy.

\subsection{Adaptation as a Prediction and Decision Problem}
These results suggest treating the adaptation procedure itself as part of the learning problem. Rather than applying one fixed fine-tuning recipe, adaptation geometry makes the update program a decision variable because alternative programs induce different tradeoffs among acquisition, transfer, boundedness, and preservation. The compiler separates two questions that are often conflated: \textit{what will happen if the model is adapted this way?} and \textit{which of those outcomes is preferred?}

This separation is useful because there need not be a universally best program. Predicting the full geometry preserves behavioral tradeoffs and allows the same predictions to be re-evaluated under different utilities without retraining. It also changes where search cost is paid. Exhaustive configuration search repeats multiple adaptations for every new episode, whereas compilation learns from prior interventions and reuses that experience. Constructing the initial geometry dataset remains expensive, but future adaptation decisions can be made from pre-adaptation information before executing only the selected program.

\subsection{What Determines an Adaptation Program?}

The experiments suggest that adaptation structure is neither purely global nor reducible to learning-objective identity. Objective-conditioned defaults are strong, but oracle programs continue to vary within objectives, and episode-conditioned prediction improves beyond those defaults on Llama. The relevant unit for adaptation therefore lies between a universal fine-tuning recipe and a categorical task-level rule.

Experiment~4 suggests that adaptation compilation may be cheaper than its most general formulation implies. Frozen representations of the learning episode alone recover nearly all of the predictive performance of the full episode--model representation, despite omitting behavioral statistics and module-level probes. In this setting, explicit diagnostics therefore appear better suited for uncertain or difficult cases than as a prerequisite for every compilation decision.


Experiment~6 adds a second source of variation: the mapping from learning problems to useful programs is model-dependent. Gemma retains clear episode-level heterogeneity even though its objective-level program preferences differ substantially from Llama. Adaptation geometry should therefore be understood as a property of the interaction among episode, model, and program rather than as a fixed localization map. Accordingly, these results should not be interpreted as evidence of strict modularity; depth and module placement are functional intervention variables, not canonical locations where particular forms of knowledge reside.

\subsection{When Should a Compiler Trust Its Prediction?}

The family- and backbone-generalization experiments expose an important boundary on the current formulation. In leave-one-family-out evaluation, performance degrades when an entire form of learning is absent from meta-training: geometry error increases, program rankings become less reliable, and compilation can underperform a fixed structural policy. The compiler therefore learns regularities over adaptation problems represented in its experience rather than a universal mapping from arbitrary learning problems to update programs.

Gemma reveals that, even when the learning families are represented and episode-level headroom remains, small errors in predicted program differences can make a deterministic argmax selector worse than a strong default. This effect is not uniform---most of the Gemma shortfall arises from lexical and especially procedural episodes, where small predicted advantages trigger program switches that do not consistently improve realized utility. Thus, successful compilation requires calibrated confidence in whether the predicted difference between competing programs is large enough to act on.

This suggests a natural extension from \emph{geometry prediction} to \emph{confidence-aware compilation}. A compiler could retain a robust global or objective-specific default when candidate programs are predicted to be nearly equivalent, and deviate only when the expected improvement is sufficiently large or well supported. Out-of-family detection, predictive uncertainty, and targeted empirical search could serve the same purpose when the compiler encounters unfamiliar adaptation problems. We do not introduce such a rule post hoc here; rather, the Gemma results identify it as an additional design requirement for future compilers.

\subsection{Practical Scope and Limitations}

Adaptation compilation is most attractive in settings involving many repeated but heterogeneous updates, where exhaustive per-episode tuning is impractical but repeatedly applying a poorly matched strategy can accumulate substantial cost. A system could maintain a library of feasible adaptation programs, characterize a new episode before training, and either execute the predicted program or fall back to a structural default or targeted search when confidence is low. Our representation ablations suggest that this decision does not always require expensive model diagnostics; richer probes could instead be reserved for uncertain cases.

The present study nevertheless uses a small and structured program space (four LoRA configurations across five controlled learning objectives). Real adaptation spaces may include finer-grained layer choices, heterogeneous rank allocation, optimizer settings, multiple adaptation mechanisms, and sequential or mixed objectives. Compilation also amortizes search rather than eliminating it: supervision still requires executing candidate programs on prior episodes, and adaptation outcomes remain stochastic. Scaling this approach will then require both richer program representations and more selective acquisition of adaptation experience.

Our broader conclusion is that adaptation itself can become learnable. Prior adaptation episodes contain information about how future learning should be carried out. Our results show that this information can support effective program selection within represented regimes, while the Gemma and leave-one-family-out experiments clarify two important limits: adaptation geometry is model-dependent, and predicted advantages must be reliable enough to justify departing from strong defaults. These boundaries turn adaptation compilation from a fixed recipe into a conditional decision problem---one that must reason not only about which program appears best, but also when that prediction is worth trusting.

\subsection*{AI use statement}

The conception of this work, including the research questions, hypotheses, methodological design, experimental planning, implementation decisions, execution of experiments, analysis, interpretation of results, and scientific conclusions, was carried out by the authors without the use of generative AI. Generative AI tools were used only as software-engineering aids to refactor portions of the existing codebase and to assist in generating and improving code documentation. All AI-assisted code changes and documentation were manually reviewed, verified against the intended functionality, and corrected where necessary by the authors.

\subsection*{Ethics statement}

This work studies methods for selecting adaptation programs for large language models based on prior adaptation experience. The experiments do not involve human subjects or the collection of personal or sensitive data. The proposed framework is intended as a methodological tool for understanding and improving how models are adapted to new learning objectives. Like other methods that improve the efficiency or precision of model adaptation, adaptation compilation is potentially dual-use: the same mechanisms that enable more targeted acquisition and preservation of desired behaviors could, in principle, be applied toward undesirable objectives. Our method does not provide guarantees regarding the safety, fairness, or downstream behavior of an adapted model, and the controlled evaluation objectives studied here should not be interpreted as such guarantees. Deployment in consequential settings would therefore require application-specific evaluation, appropriate safeguards, and consideration of the properties and limitations of the underlying model.


\subsection*{Reproducibility statement}
We provide the materials needed to reproduce the experimental pipeline and reported analyses. Section~\ref{sec:methods} specifies the learning episodes, data splits, adaptation-program space, utility functions, optimization calibration, adaptation records, pre-adaptation features, and predictor-selection protocol; Sections~\ref{sec:exp1}--\ref{sec:exp6} define the evaluation procedures for each experiment. Appendices~\ref{app:calibration}--\ref{app:gemma} provide additional calibration results, analyses of selection headroom and adaptation stochasticity, geometry-prediction and program-selection results, representation ablations, leave-one-family-out evaluation, and the complete Gemma replication protocol. The supplementary materials, including code, datasets, experimental outputs, and configuration files, are available in our \href{https://github.com/rramnauth2220/adaptcompile-experiments}{GitHub repository}. A reusable implementation of adaptation compilation is also available as the \href{https://pypi.org/project/adaptcompile/}{\texttt{adaptcompile} Python package}. Model and hyperparameter selection use only meta-training and validation data, and all reported test results use fixed held-out episodes and the adaptation seeds specified in the experimental protocol.

\subsubsection*{Author Contributions}
Rebecca Ramnauth conceived the project, developed the adaptation compilation framework, designed and implemented the experimental methodology, conducted the experiments, analyzed the results, and led the writing of the manuscript. Brian Scassellati contributed to manuscript revision. All authors reviewed and approved the final manuscript.

\bibliography{references}
\bibliographystyle{iclr2027_conference}

\appendix
\section*{Appendices}
The appendices provide additional methodological detail and analyses supporting the main experiments. We first document optimization calibration (Appendix~\ref{app:calibration}), then expand the Llama results with analyses of selection headroom and stochasticity, geometry prediction, program selection, and representation ablations (Appendices~\ref{app:exp1}--\ref{app:exp4}). We next examine generalization to unseen learning families (Appendix~\ref{app:exp5}) and conclude with the full Gemma replication protocol and results (Appendix~\ref{app:gemma}).

\section{Optimization Schedule Calibration}
\label{app:calibration}

Because each adaptation episode contains a single latent specification, we calibrate the optimization schedule separately for this single-specification regime before comparing adaptation configurations. The purpose of this calibration is to ensure that subsequent differences in adaptation geometry are not artifacts of systematic undertraining.

We use five meta-training episodes from each of the five learning objectives ($25$ episodes total) and adapt each episode using the unconstrained full-stack LoRA configuration. Holding the remaining optimization settings fixed, we vary gradient accumulation over $\{1,2,4,8\}$, thereby varying the number of optimizer updates available during adaptation. No validation or test episodes are used for calibration.

\begin{table}[h]
\centering
\caption{Optimization calibration on meta-training episodes. Values are averaged across 25 episodes.}
\label{tab:optimization-calibration}
\begin{tabular}{ccccc}
\toprule
Gradient Accum. & Acquisition & Transfer & Boundedness & Preservation \\
\midrule
8 & 0.408 & 0.240 & 0.393 & 0.997 \\
4 & 0.648 & 0.587 & 0.500 & 0.992 \\
2 & 0.896 & \textbf{0.867} & 0.547 & 0.971 \\
1 & \textbf{0.904} & 0.853 & \textbf{0.700} & 0.876 \\
\bottomrule
\end{tabular}
\end{table}

Larger accumulation values leave several learning objectives near the acquisition or transfer floor. Reducing gradient accumulation to $2$ substantially increases both acquisition and transfer across objectives. A further reduction to $1$ yields negligible additional acquisition ($0.896\rightarrow0.904$) and slightly lower transfer ($0.867\rightarrow0.853$), while preservation decreases substantially ($0.971\rightarrow0.876$). We therefore select gradient accumulation $2$ as the least aggressive schedule that reliably supports learning across objectives without unnecessary collateral interference. This optimization schedule is fixed for all subsequent configurations, episodes, and evaluation splits.

\section{Experiment 1: Characterizing Selection Headroom}
\label{app:exp1}

The main text establishes that alternative adaptation programs induce meaningful selection headroom on held-out episodes. Here, we provide the full objective-level results, characterize the separation between oracle-optimal programs, and examine robustness to adaptation stochasticity. All headline test results use the same 100 held-out episodes reported in the main text, with each configuration's behavioral outcomes averaged across three adaptation seeds.

\subsection{Selection Headroom by Learning Objective}

Table~\ref{tab:exp1_headroom} reports the full selection results by learning objective. The global-fixed program is selected using all meta-training episodes, while the objective-fixed policy selects one program per learning objective using only meta-training episodes. The episode oracle selects the highest-utility program independently for each held-out episode.

\begin{table}[h]
    \centering
    \small
    \setlength{\tabcolsep}{4pt}
    \caption{
    \textbf{Selection headroom by learning objective.}
    Utilities are measured on held-out test episodes. Global- and objective-fixed programs are selected using meta-training data only. Regret is measured relative to the per-episode oracle, and ``Opt.'' denotes the fraction of episodes on which the corresponding fixed policy is oracle-optimal.
    }
    \label{tab:exp1_headroom}
    \resizebox{\linewidth}{!}{
    \begin{tabular}{llrrrrrrr}
        \toprule
        Objective &
        Objective-fixed &
        Global $U$ &
        Obj. $U$ &
        Oracle $U$ &
        Global Regret &
        Obj. Regret &
        Global Opt. &
        Obj. Opt. \\
        \midrule
        Overall
            & ---    & .569 & .600 & .618 & .049 & .018 & .41 & .64 \\
        Behavioral
            & Late   & .714 & .747 & .753 & .039 & .007 & .15 & .45 \\
        Causal
            & Middle & .489 & .489 & .507 & .018 & .018 & .75 & .75 \\
        Factual
            & Late   & .504 & .564 & .591 & .087 & .027 & .10 & .40 \\
        Lexical
            & Early  & .688 & .751 & .790 & .102 & .039 & .10 & .65 \\
        Procedural
            & Middle & .449 & .449 & .450 & .001 & .001 & .95 & .95 \\
        \bottomrule
    \end{tabular}
    }
\end{table}

Objective conditioning accounts for a substantial portion of program preference, reducing mean regret from $0.049$ under the global-fixed policy to $0.018$. However, the magnitude of the remaining episode-specific headroom differs considerably across learning objectives. Lexical binding and factual association retain the largest objective-to-oracle gaps, whereas procedural reasoning admits a nearly universal middle-depth default. In the latter case, the objective-fixed policy is oracle-optimal on $95\%$ of held-out episodes and differs from the oracle by less than $0.001$ utility on average.

\subsection{Oracle Programs and Winner Separation}

Table~\ref{tab:exp1_winners} gives the distribution of oracle-optimal programs underlying Fig.~\ref{fig:exp1} in the main text. We assign fractional credit when multiple configurations attain the same maximum utility. Across all 100 test episodes, middle-depth adaptation is oracle-optimal most frequently, but every candidate program is preferred for a nontrivial subset of episodes.

\begin{table}[h]
    \centering
    \small
    \setlength{\tabcolsep}{4pt}
    \caption{
    \textbf{Oracle-program distribution and winner separation.}
    Program columns report the fraction of held-out episodes for which each configuration is oracle-optimal, using fractional credit for ties. The final columns report the tie rate and the mean and median utility margin between the highest- and second-highest-utility configurations.
    }
    \label{tab:exp1_winners}
    \begin{tabular}{lrrrrrrr}
        \toprule
        Objective &
        Early &
        Middle &
        Late &
        Full-r4 &
        Tie Rate &
        Mean Margin &
        Median Margin \\
        \midrule
        Overall
            & .165 & .410 & .210 & .215 & .01 & .054 & .041 \\
        Behavioral
            & .000 & .150 & .450 & .400 & .00 & .016 & .007 \\
        Causal
            & .000 & .750 & .150 & .100 & .00 & .072 & .059 \\
        Factual
            & .175 & .100 & .400 & .325 & .05 & .045 & .023 \\
        Lexical
            & .650 & .100 & .000 & .250 & .00 & .077 & .076 \\
        Procedural
            & .000 & .950 & .050 & .000 & .00 & .059 & .053 \\
        \bottomrule
    \end{tabular}
\end{table}

\begin{figure}[h]
    \centering
    \includegraphics[width=0.6\linewidth]{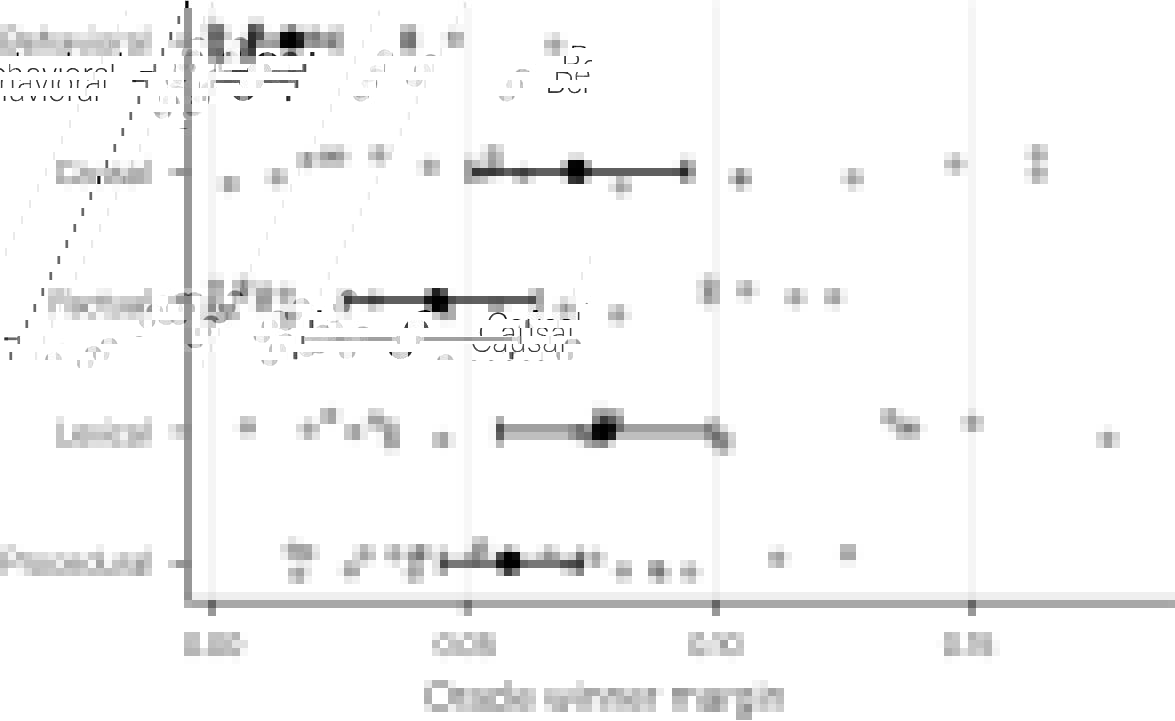}
    \caption{
    \textbf{Separation between oracle-optimal and second-best programs.}
    Points show the utility margin between the best and second-best adaptation programs for individual held-out episodes, grouped by learning objective. Summary markers indicate the objective-level mean.
    }
    \label{fig:exp1_margins}
\end{figure}

The variation in oracle programs is not primarily driven by ties. Only $1\%$ of test episodes contain multiple utility-maximizing configurations, and $99\%$ therefore have a unique winner. The mean difference between the best and second-best configurations is $0.054$ utility (median $0.041$). Fig.~\ref{fig:exp1_margins} shows these margins at the episode level.

\subsection{Robustness to Adaptation Stochasticity}

\begin{figure}[h]
    \centering
    \includegraphics[width=\linewidth]{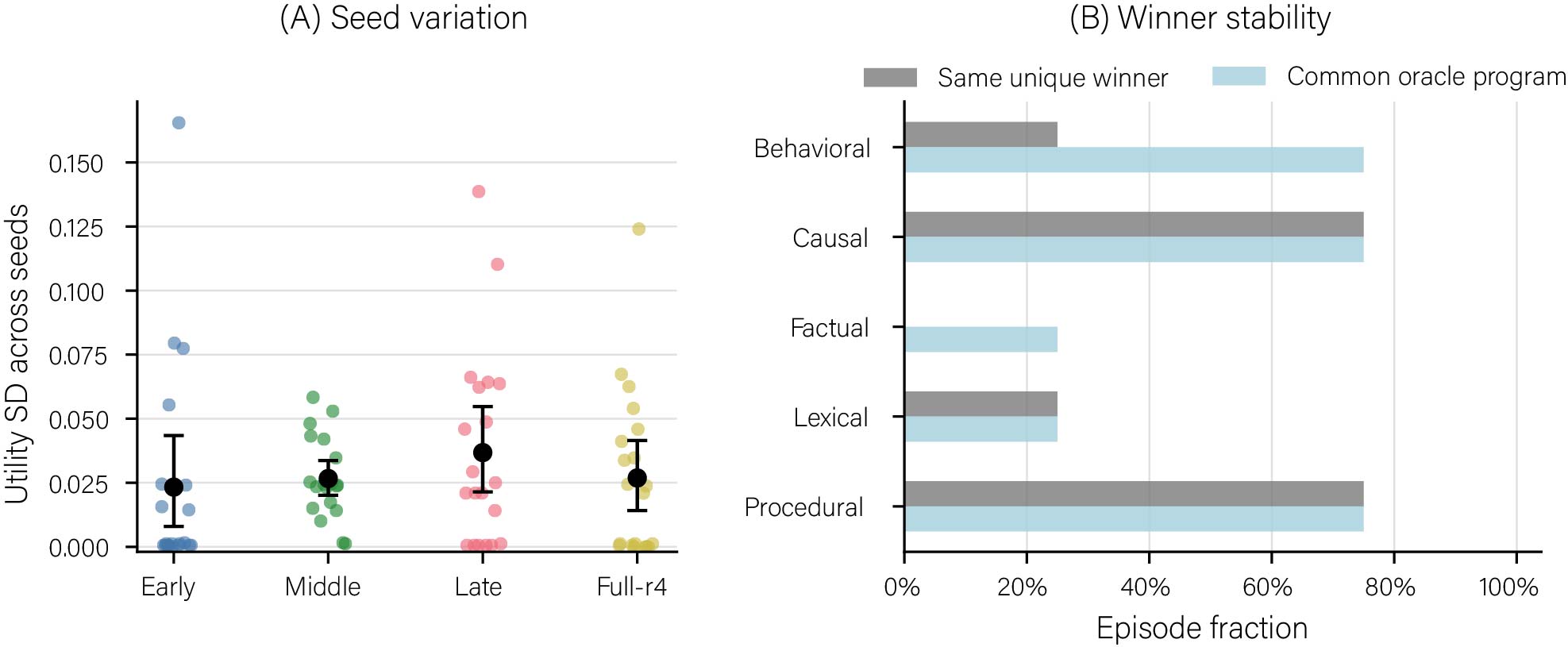}
    \caption{
    \textbf{Adaptation stochasticity across programs.}
    \textbf{(A)} Variation in utility across three adaptation seeds for each episode--program pair.
    \textbf{(B)} Stability of the oracle-optimal program across seeds, using both strict unique-winner and tie-aware criteria.
    }
    \label{fig:exp1_seed_robustness}
\end{figure}

Adaptation outcomes also vary across optimization runs. To characterize this variability directly, we analyze a stratified subset of 20 held-out episodes for which all four candidate programs were independently adapted under three random seeds. Our headline test geometry averages three seeds for all 100 held-out episodes. Here, for each episode--configuration pair, we measure the variation in utility across seeds and examine whether the identity of the oracle-optimal program remains stable (Fig.~\ref{fig:exp1_seed_robustness}).

Adaptation stochasticity is non-negligible: the mean within-configuration standard deviation in utility is $0.028$, and only $40\%$ of episodes retain the same unique oracle winner across all three individual seeds. A less restrictive, tie-aware criterion finds at least one common oracle configuration across all seeds for $55\%$ of episodes. These results motivate defining adaptation geometry in terms of expected behavioral outcomes rather than the realization of a single optimization run.

Importantly, selection headroom persists after averaging over this stochasticity. On the robustness subset, the global-fixed, objective-fixed, and episode-oracle utilities are $0.581$, $0.605$, and $0.621$, respectively. The resulting objective-to-oracle regret remains $0.016$, and the objective-fixed policy is oracle-optimal on only $60\%$ of episodes.

\begin{figure}[h]
    \centering
    \includegraphics[width=\linewidth]{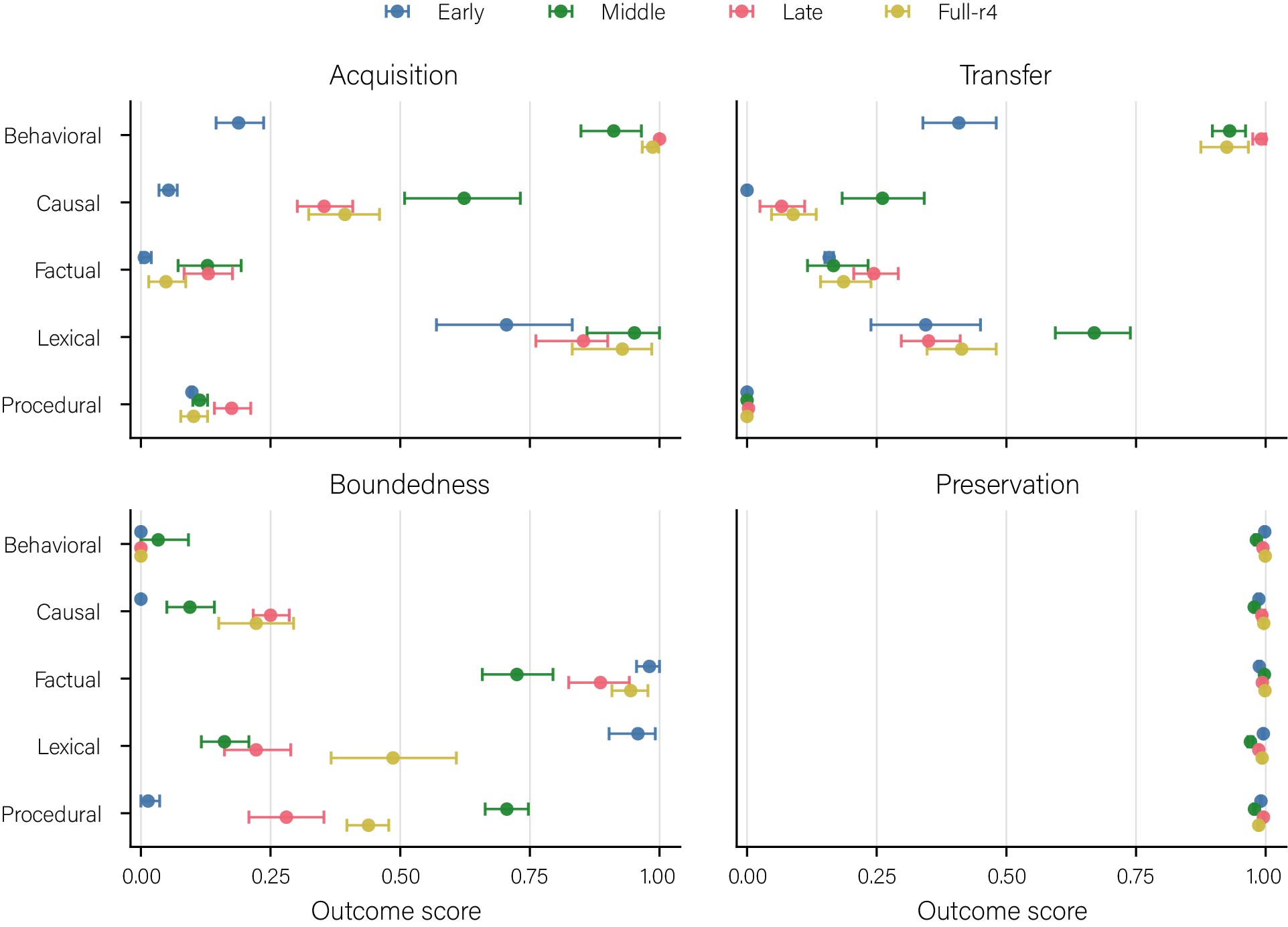}
    \caption{
    \textbf{Configuration sensitivity across adaptation outcomes.}
    Acquisition, transfer, boundedness, and preservation under each budget-matched adaptation program, grouped by learning objective. Values summarize held-out test episodes using seed-averaged outcomes.
    }
    \label{fig:exp1_outcomes}
\end{figure}

\subsection{Configuration Sensitivity Across Behavioral Outcomes}

Selection utility compresses four distinct behavioral criteria into a single scalar. To verify that program sensitivity is not restricted to aggregate utility, Fig.~\ref{fig:exp1_outcomes} decomposes adaptation outcomes into acquisition, transfer, boundedness, and preservation. Alternative programs induce distinct behavioral tradeoffs across these dimensions, providing the outcome-level variation from which the selection differences reported in the main text arise.

\section{Experiment 2: Predicting Adaptation Geometry}
\label{app:exp2}

We select the predictor using validation mean absolute error averaged across acquisition, transfer, boundedness, and preservation. The candidate models are ridge regression and random-forest regression; validation selects the random forest ($0.044$ MAE, compared with $0.099$ for ridge). The selected predictor uses 300 trees, unrestricted depth, a minimum leaf size of one, and random seed 2026. Model selection and hyperparameter choice use only meta-training and validation episodes.

\begin{figure}[t]
    \centering
    \includegraphics[width=0.9\linewidth]{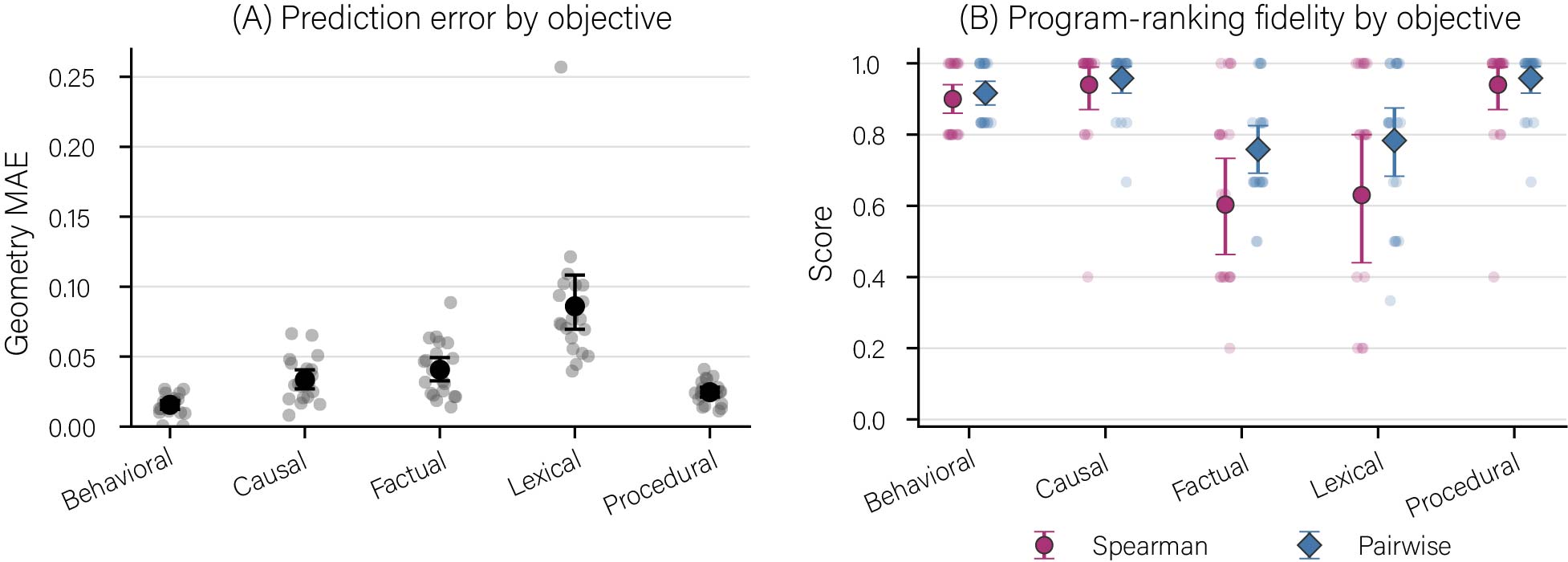}
    \caption{
    \textbf{Geometry prediction across learning objectives.}
    \textbf{(A)} Episode-level mean absolute error between predicted and observed adaptation geometry, grouped by learning objective. Gray points denote held-out episodes and black markers indicate means with 95\% bootstrap confidence intervals.
    \textbf{(B)} Within-episode agreement between predicted and observed program orderings, measured by Spearman rank correlation and pairwise ranking accuracy. Points denote individual episodes and summary markers indicate means with 95\% bootstrap confidence intervals. The same predictor is evaluated across all objectives without access to objective identity.
    }
    \label{fig:exp2_by_objective}
\end{figure}

The aggregate prediction results in Sec.~\ref{sec:exp2} obscure substantial variation across learning objectives. Table~\ref{tab:exp2_by_objective}, depicted as Fig.~\ref{fig:exp2_by_objective}, therefore evaluates the same frozen geometry predictor separately within each objective. Prediction error is lowest for behavioral episodes and remains relatively low for causal, factual, and procedural episodes, while lexical binding is markedly more variable and contains the largest-error episodes. The same distinction appears in program-ranking fidelity: predicted geometry preserves program order particularly well for behavioral, causal, and procedural episodes, whereas factual and lexical episodes are more difficult to rank. Thus, the aggregate performance of the predictor is not attributable to a single learning family, although the reliability of episode-level geometry prediction differs across objectives.

\begin{table}[h]
    \centering
    \small
    \setlength{\tabcolsep}{4pt}
    \caption{
    \textbf{Geometry prediction by learning objective.}
    Geometry MAE is averaged across acquisition, transfer, boundedness, and preservation. Spearman correlation and pairwise accuracy measure agreement between predicted and observed program rankings within each episode. The same primary predictor is evaluated across all objectives without access to objective identity.
    }
    \label{tab:exp2_by_objective}
    \resizebox{\linewidth}{!}{
    \begin{tabular}{lrrrrrrr}
        \toprule
        Objective &
        Geometry MAE &
        Acquisition &
        Transfer &
        Boundedness &
        Preservation &
        Spearman &
        Pairwise \\
        \midrule
        Behavioral
            & .015 & .023 & .030 & .007 & .002 & .900 & .917 \\
        Causal
            & .033 & .050 & .051 & .030 & .002 & .940 & .958 \\
        Factual
            & .041 & .045 & .053 & .061 & .003 & .603 & .758 \\
        Lexical
            & .086 & .090 & .138 & .112 & .004 & .630 & .783 \\
        Procedural
            & .025 & .022 & .001 & .073 & .002 & .940 & .958 \\
        \bottomrule
    \end{tabular}
    }
\end{table}

\section{Experiment 3: Evaluating Program Selection}
\label{app:exp3}
These analyses clarify where compilation improves on structural defaults and how reusable the predicted geometry remains under changing behavioral priorities.

\subsection{Selection by learning objective}
\begin{figure}[t]
    \centering
    \includegraphics[width=0.9\linewidth]{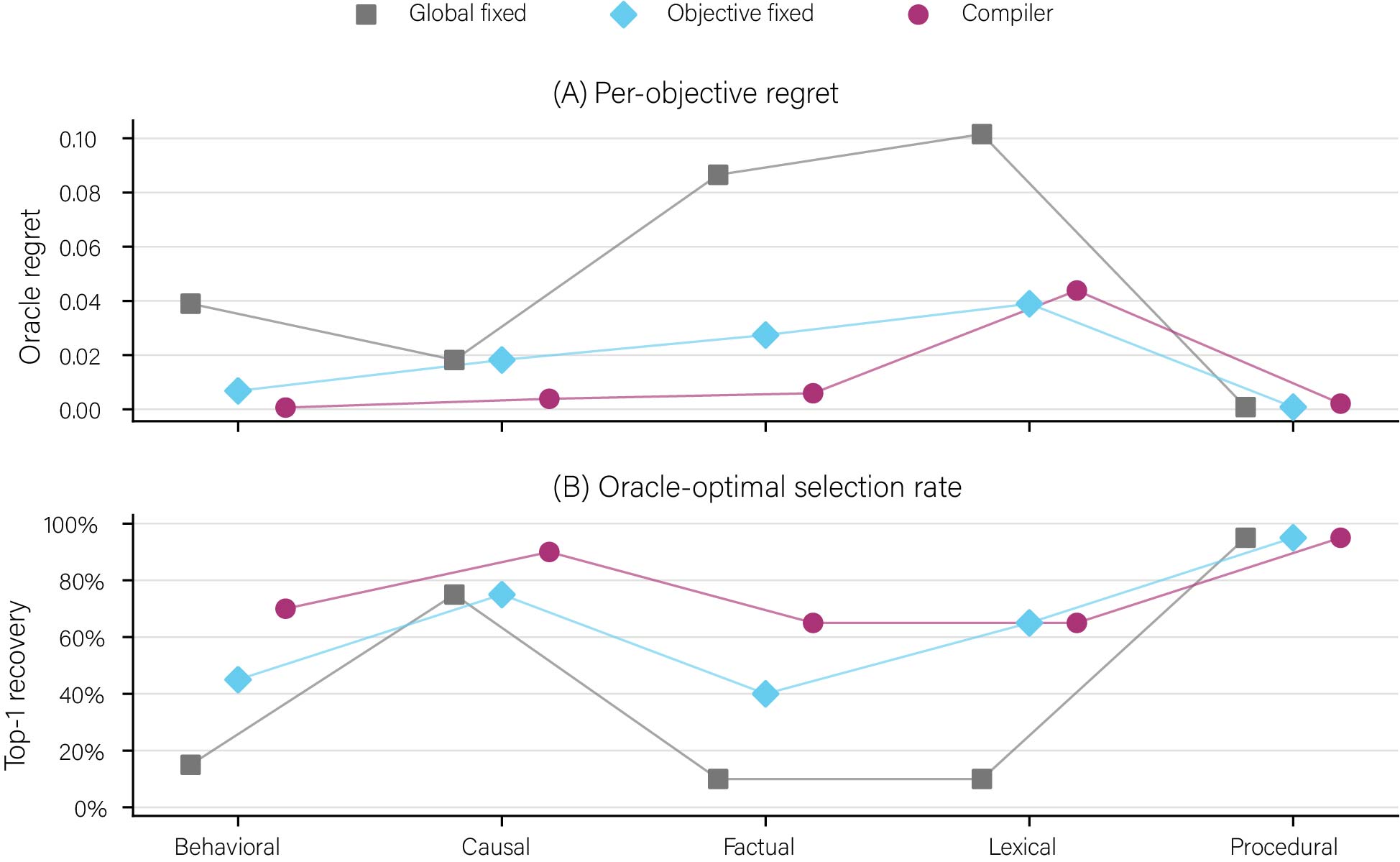}
    \caption{
    \textbf{Program-selection performance by learning objective.}
    \textbf{(A)} Oracle regret under global-fixed, objective-fixed, and compiler selection.
    \textbf{(B)} Fraction of held-out episodes for which each policy selects an oracle-optimal program.
    Compiler performance is strongest for behavioral, causal, factual, and procedural episodes, while lexical binding remains the principal residual failure case.
    }
    \label{fig:exp3_by_objective}
\end{figure}

Table~\ref{tab:exp3_by_objective}, depicted as Fig.~\ref{fig:exp3_by_objective}, decomposes balanced-utility selection performance by learning objective. The compiler substantially reduces oracle regret relative to the global-fixed policy across behavioral, causal, factual, and lexical episodes, and approaches the oracle particularly closely for behavioral and causal learning. Procedural reasoning already admits a strong fixed structural default, leaving little headroom for episode-conditioned selection.

\begin{table}[t]
    \centering
    \small
    \setlength{\tabcolsep}{4pt}
    \caption{
    \textbf{Program selection by learning objective.}
    Realized utility and oracle regret are reported for the global-fixed, objective-fixed, and compiler policies. Top-1 denotes the fraction of held-out episodes for which each policy selects an oracle-optimal program.
    }
    \label{tab:exp3_by_objective}
    \begin{tabular}{lrrrrrrrrr}
        \toprule
        & \multicolumn{3}{c}{Realized Utility}
        & \multicolumn{3}{c}{Oracle Regret}
        & \multicolumn{3}{c}{Top-1} \\
        \cmidrule(lr){2-4}
        \cmidrule(lr){5-7}
        \cmidrule(lr){8-10}
        Objective &
        Global & Obj. & Compiler &
        Global & Obj. & Compiler &
        Global & Obj. & Compiler \\
        \midrule
        Behavioral
            & .714 & .747 & .753
            & .039 & .007 & .001
            & .15 & .45 & .70 \\
        Causal
            & .489 & .489 & .504
            & .018 & .018 & .004
            & .75 & .75 & .90 \\
        Factual
            & .504 & .564 & .585
            & .087 & .027 & .006
            & .10 & .40 & .65 \\
        Lexical
            & .688 & .751 & .746
            & .102 & .039 & .044
            & .10 & .65 & .65 \\
        Procedural
            & .449 & .449 & .448
            & .001 & .001 & .002
            & .95 & .95 & .95 \\
        \bottomrule
    \end{tabular}
\end{table}

Lexical binding is the clearest exception to the general improvement over the objective-fixed policy. Although the compiler selects an oracle-optimal program on $65\%$ of lexical episodes, matching the objective-fixed policy, its mean realized utility is slightly lower ($0.746$ versus $0.751$) because its mistakes occur on higher-regret episodes. This is consistent with the greater lexical prediction error observed in Appendix~\ref{app:exp2}. By contrast, factual association retains substantial objective-level headroom and benefits strongly from episode-conditioned selection, with regret decreasing from $0.027$ to $0.006$.

\subsection{Paired comparison with objective-fixed selection}

Across all 100 held-out episodes, the compiler improves realized utility over the objective-fixed policy on 22 episodes, performs worse on 7, and selects a program with identical realized utility on the remaining 71. The mean paired gain is $0.0072$ utility; an episode-level bootstrap gives a 95\% confidence interval of $[0.0011,\,0.0136]$. Thus, the aggregate improvement over the objective-fixed baseline is not produced by symmetric exchanges of similarly valued programs: departures from the objective-level default are more often beneficial than harmful.

\subsection{Recompilation under alternative utility specifications}

Because the predictor estimates the components of adaptation geometry rather than a single preferred program, changing the utility function requires only rerunning program selection. Table~\ref{tab:exp3_utility_specs} reports the resulting performance and the fraction of compiler decisions that change relative to balanced utility.

\begin{table}[t]
    \centering
    \small
    \setlength{\tabcolsep}{5pt}
    \caption{
    \textbf{Recompilation under alternative utility specifications.}
    The same predicted adaptation geometry is evaluated under each utility without retraining the predictor. Regret reduction is measured relative to the objective-fixed policy. ``Switch'' reports the fraction of compiler-selected programs that differ from the balanced-utility selection.
    }
    \label{tab:exp3_utility_specs}
    \resizebox{\linewidth}{!}{
    \begin{tabular}{lrrrrrr}
        \toprule
        Utility &
        Compiler $U$ &
        Oracle $U$ &
        Compiler Regret &
        Obj.-fixed Regret &
        Regret Reduction &
        Switch \\
        \midrule
        Balanced
            & .607 & .618 & .011 & .018 & 38.9\% & --- \\
        Transfer-heavy
            & .567 & .578 & .011 & .025 & 53.4\% & 14\% \\
        Boundedness-heavy
            & .602 & .606 & .004 & .025 & 84.2\% & 13\% \\
        Preservation-heavy
            & .683 & .692 & .009 & .015 & 39.1\% & 1\% \\
        \bottomrule
    \end{tabular}
    }
\end{table}

Reweighting transfer or boundedness changes the selected program for approximately $14\%$ and $13\%$ of episodes, respectively, whereas increasing the weight on preservation changes only $1\%$ of selections. This difference is consistent with preservation being both comparatively stable across candidate programs and accurately predicted in Experiment~2. More importantly, the compiler remains closer to the oracle than the objective-fixed policy under every utility specification, without retraining the geometry predictor.

\section{Experiment 4: Ablating Pre-Adaptation Information}
\label{app:exp4}

\begin{figure}[h]
    \centering
    \includegraphics[width=\linewidth]{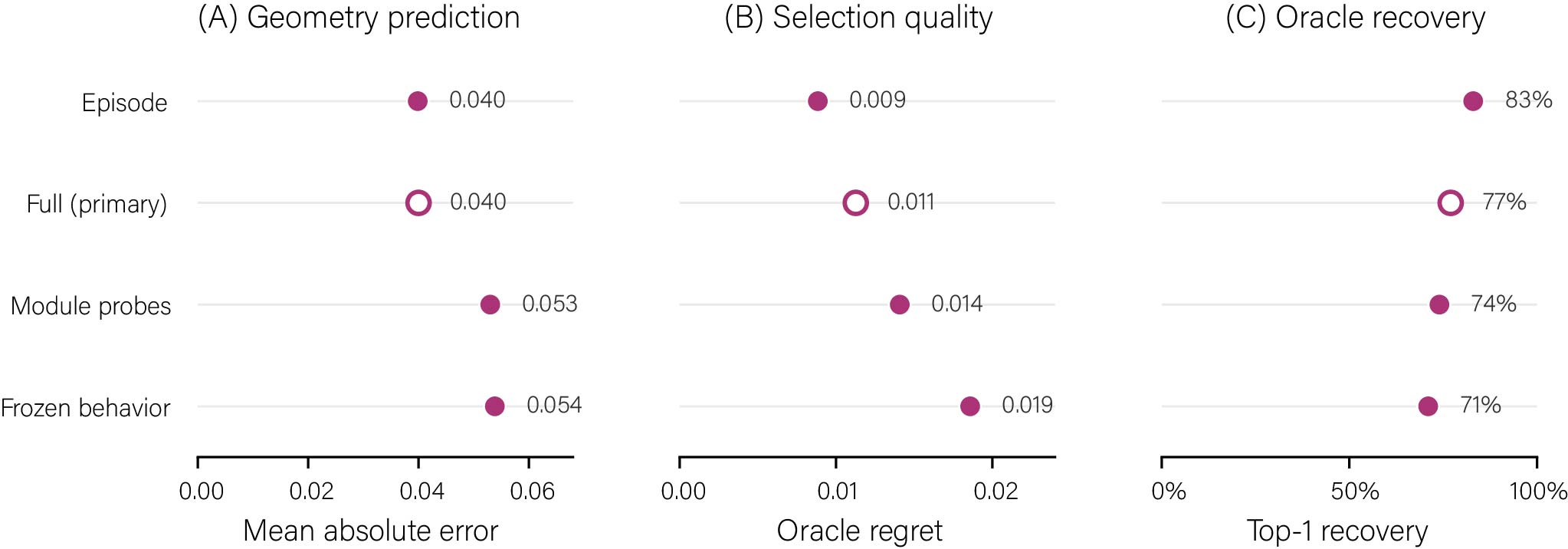}
    \caption{
    \textbf{Representation ablations.}
    We compare the full episode--model representation with three restricted feature sets.
    \textbf{(A)} Mean absolute error in predicted adaptation geometry.
    \textbf{(B)} Oracle regret of the program selected from predicted geometry.
    \textbf{(C)} Tie-aware top-1 recovery of an oracle-optimal program.
    The episode representation alone retains performance comparable to the full representation, while module-level probes and frozen-model behavioral statistics are weaker when used independently. The outlined marker denotes the full representation used as the primary predictor in Experiments~2 and 3.
    }
    \label{fig:exp4}
\end{figure}

Table~\ref{tab:exp4_full_ablation} reports the complete prediction and selection metrics for each representation condition. These data are summarized and contextualized alongside the results of Experiments 2 and 3 in Fig.~\ref{fig:exp4}.  Table~\ref{tab:exp4_features} summarizes the information available to each predictor. All variants use the same candidate-program descriptors and are selected using validation episodes before evaluation on the held-out test set.

\begin{table}[h]
    \centering
    \small
    \setlength{\tabcolsep}{4pt}
    \caption{
    \textbf{Complete representation-ablation results.}
    Each restricted predictor is selected using validation performance and evaluated on the same held-out test episodes. Geometry MAE measures prediction error across acquisition, transfer, boundedness, and preservation. Ranking metrics are computed across candidate programs within each episode.
    }
    \label{tab:exp4_full_ablation}
    \resizebox{\linewidth}{!}{
    \begin{tabular}{lrrrrrrrr}
        \toprule
        Representation &
        Val. MAE &
        Test MAE &
        Utility Corr. &
        Spearman &
        Pairwise &
        Top-1 &
        Top-2 &
        Oracle Regret \\
        \midrule
        Episode
            & .0434 & .0399 & .9633 & .8478 & .8983 & .83 & .96 & .0088 \\
        Full
            & .0435 & .0400 & .9620 & .8027 & .8750 & .77 & .93 & .0113 \\
        Module probes
            & .0526 & .0530 & .9348 & .7655 & .8600 & .74 & .89 & .0141 \\
        Frozen behavior
            & .0563 & .0538 & .9176 & .7406 & .8400 & .71 & .91 & .0186 \\
        \bottomrule
    \end{tabular}
    }
\end{table}

\begin{table}[t]
    \centering
    \small
    \caption{
    \textbf{Feature sets used in the representation ablation.}
    Configuration descriptors are provided to all predictors; rows describe the episode--model information available in each condition.
    }
    \label{tab:exp4_features}
    \resizebox{\linewidth}{!}{
    \begin{tabular}{lcccc}
        \toprule
        Representation &
        Hidden-state representation &
        Frozen-model loss &
        Module diagnostics &
        Backward pass required \\
        \midrule
        Episode
            & \checkmark & -- & -- & No \\
        Frozen behavior
            & -- & \checkmark & -- & No \\
        Module probes
            & -- & -- & \checkmark & Yes \\
        Full
            & \checkmark & \checkmark & \checkmark & Yes \\
        \bottomrule
    \end{tabular}
    }
\end{table}

\begin{table}[t]
    \centering
    \small
    \setlength{\tabcolsep}{3.5pt}
    \caption{
    \textbf{Representation ablations by learning objective.}
    Geometry MAE measures prediction error across acquisition, transfer,
    boundedness, and preservation; oracle regret measures the downstream
    quality of the program selected from each predicted geometry.
    Bold values indicate the best result within each objective and metric group.
    }
    \label{tab:exp4_by_objective}
    \begin{tabular}{lrrrrrrrr}
        \toprule
        &
        \multicolumn{4}{c}{Geometry MAE $\downarrow$} &
        \multicolumn{4}{c}{Oracle Regret $\downarrow$} \\
        \cmidrule(lr){2-5}
        \cmidrule(lr){6-9}
        Objective &
        Episode & Full & Probe & Frozen &
        Episode & Full & Probe & Frozen \\
        \midrule
        Behavioral
            & \textbf{.0153} & .0155 & .0155 & .0167
            & \textbf{.0005} & .0006 & \textbf{.0005} & .0009 \\

        Causal
            & .0346 & .0335 & .0450 & \textbf{.0323}
            & \textbf{.0039} & \textbf{.0039} & .0092 & .0074 \\

        Factual
            & \textbf{.0379} & .0406 & .0679 & .0501
            & \textbf{.0007} & .0059 & .0236 & .0345 \\

        Lexical
            & .0864 & \textbf{.0860} & .1100 & .1062
            & \textbf{.0370} & .0438 & \textbf{.0370} & .0423 \\

        Procedural
            & .0252 & \textbf{.0245} & .0267 & .0639
            & .0021 & .0021 & \textbf{.0000} & .0078 \\
        \bottomrule
    \end{tabular}
\end{table}

Table~\ref{tab:exp4_by_objective} decomposes the representation ablations by
learning objective. No restricted representation dominates every objective:
episode representations perform particularly well for behavioral and factual
episodes, while full representations yield the lowest geometry error for
lexical and procedural episodes. Prediction error and selection quality also
need not coincide; for example, module probes achieve zero oracle regret on
procedural episodes despite not minimizing geometry MAE.

\section{Experiment 5: Generalizing to Unseen Learning Families}
\label{app:exp5}

For each leave-one-family-out (LOFO) evaluation, the indicated learning objective is excluded entirely from both meta-training and validation. The predictor is trained on 320 episodes and selected using 80 validation episodes from the remaining four objectives, then evaluated on the 20 test episodes from the held-out objective.

\begin{figure}[h]
    \centering
    \includegraphics[width=0.85\linewidth]{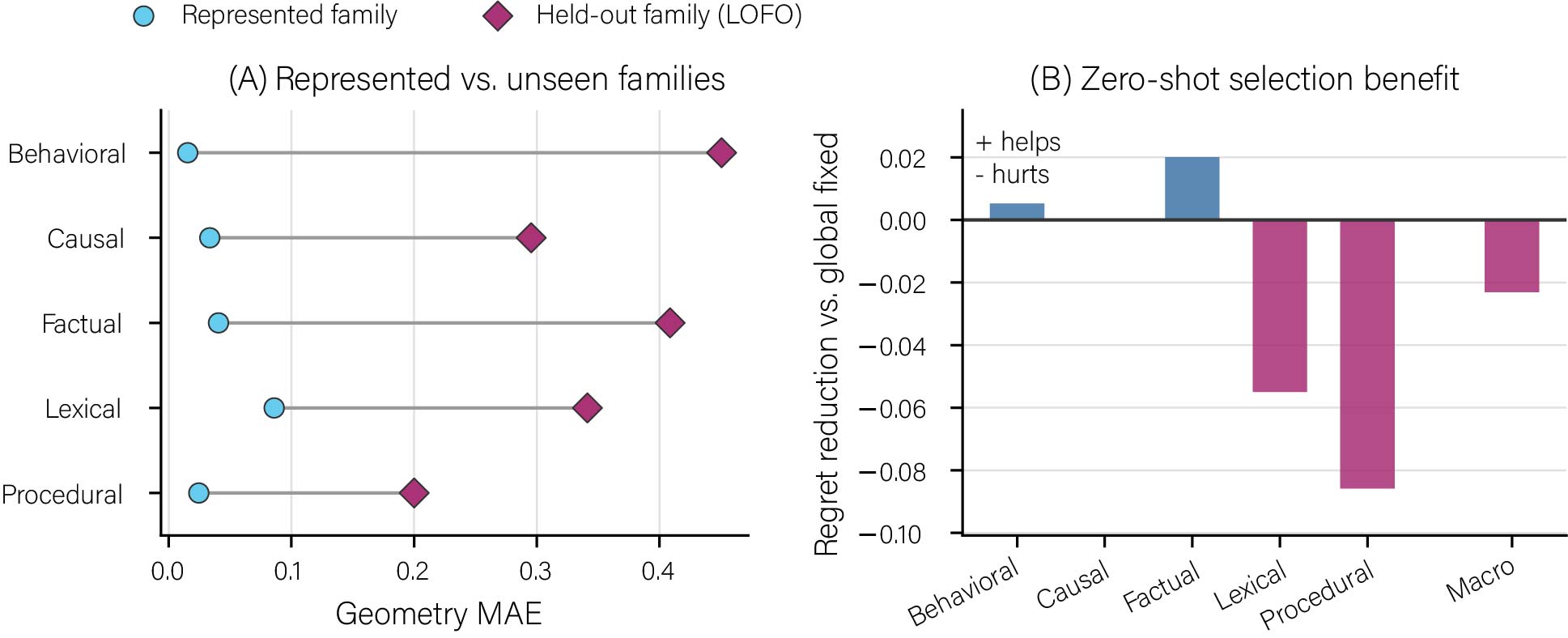}
    \caption{
    \textbf{Generalization beyond represented learning families.}
    \textbf{(A)} Geometry prediction error for held-out episodes from learning families represented during meta-training compared with leave-one-family-out (LOFO) evaluation, where the test objective is excluded from both training and validation.
    \textbf{(B)} Change in oracle regret of the LOFO compiler relative to the global-fixed policy; positive values indicate improved selection and negative values worse selection.
    }
    \label{fig:exp5_lofo}
\end{figure}

\begin{figure}[h]
    \centering
    \includegraphics[width=0.75\linewidth]{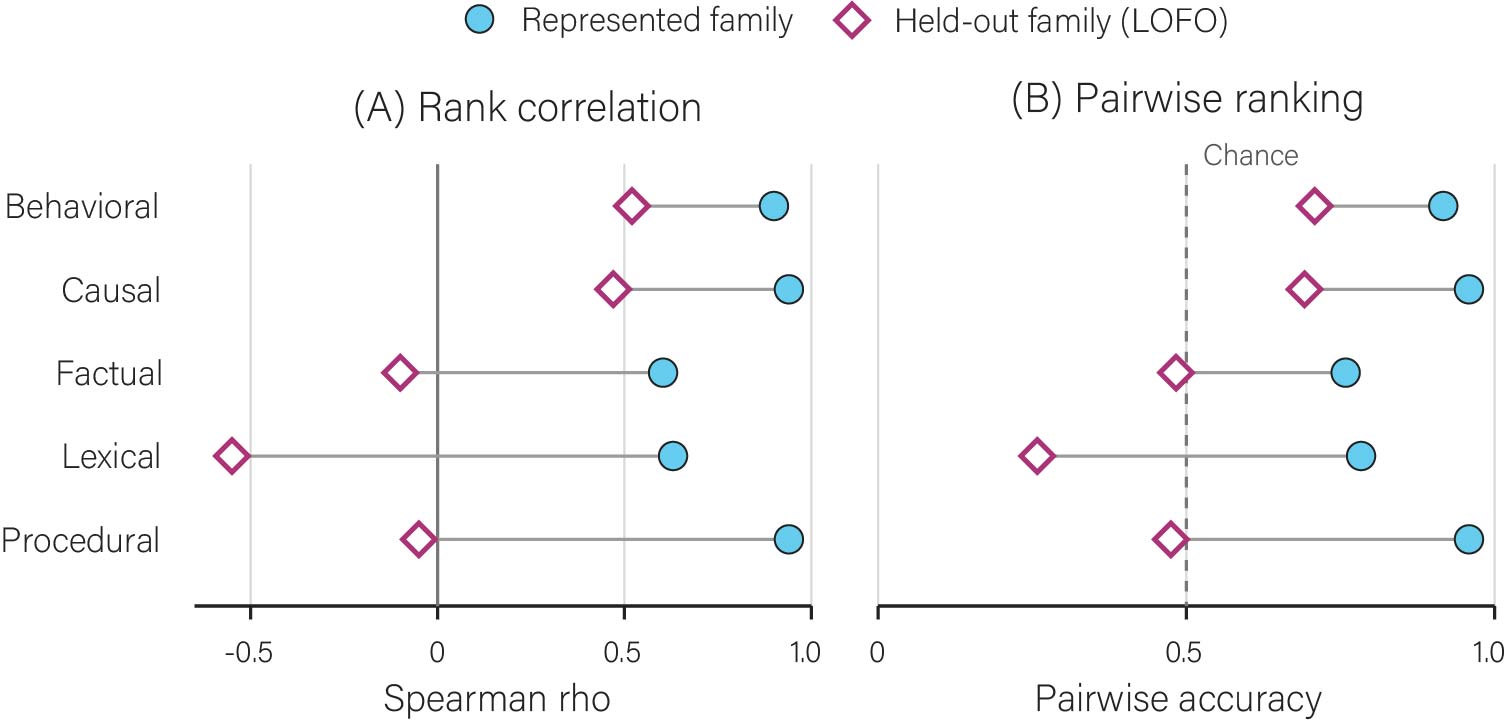}
    \caption{
    \textbf{Program-ranking fidelity under leave-one-family-out generalization.}
    Within-episode agreement between predicted and observed adaptation-program
orderings is compared when the learning family is represented during
meta-training versus excluded entirely under LOFO
evaluation. (\textbf{A}) Spearman rank correlation. (\textbf{B}) Pairwise ranking accuracy.
Ranking fidelity decreases across all five objectives when the learning family
is unseen, with particularly large degradation for factual, lexical, and
procedural episodes.
    }
    \label{fig:exp5_ranking}
\end{figure}

Fig.~\ref{fig:exp5_lofo} summarizes the resulting distribution shift.
Relative to held-out episodes from represented learning families, excluding an entire family substantially increases geometry-prediction error and degrades program selection. Table~\ref{tab:exp5_lofo_prediction} decomposes prediction and ranking performance by held-out family. Fig.~\ref{fig:exp5_ranking} shows that within-episode ranking fidelity declines across all five held-out families.

Table~\ref{tab:exp5_lofo_selection} shows that this degradation is heterogeneous. Zero-shot compilation reduces regret for behavioral and factual episodes, is neutral for causal mapping, and increases regret for lexical and procedural episodes. At the macro level, compiler regret increases from $.062$ under the global-fixed policy to $.085$ under LOFO selection.

\begin{table}[t]
    \centering
    \small
    \setlength{\tabcolsep}{4pt}
    \caption{
    \textbf{Leave-one-family-out geometry prediction.}
    For each fold, the indicated learning family is excluded entirely from both meta-training and validation. Validation MAE is measured on represented learning families, while test metrics are computed on episodes from the held-out family. Ranking metrics compare predicted and observed program utilities within each episode.
    }
    \label{tab:exp5_lofo_prediction}
    \begin{tabular}{lrrrrrrr}
        \toprule
        Held-out family &
        Val. MAE &
        Test MAE &
        Utility Corr. &
        Spearman &
        Pairwise &
        Top-1 &
        Top-2 \\
        \midrule
        Behavioral
            & .050 & .451 & .862  & .520  & .708 & .30 & .60 \\
        Causal
            & .044 & .295 & .646  & .470  & .692 & .10 & .55 \\
        Factual
            & .043 & .409 & -.110 & -.100 & .483 & .20 & .45 \\
        Lexical
            & .026 & .341 & -.026 & -.550 & .258 & .05 & .15 \\
        Procedural
            & .046 & .200 & -.037 & -.050 & .475 & .05 & .60 \\
        \midrule
        Macro
            & .042 & .339 & .267  & .058  & .523 & .14 & .47 \\
        \bottomrule
    \end{tabular}
\end{table}

\begin{table}[h]
    \centering
    \small
    \setlength{\tabcolsep}{4pt}
    \caption{
    \textbf{Program selection under leave-one-family-out generalization.}
    Utilities are realized under balanced utility. Regret is measured relative
    to the exhaustive episode oracle. $\Delta R$ denotes regret reduction
    relative to the global-fixed policy, such that positive values indicate
    beneficial zero-shot compilation.
    }
    \label{tab:exp5_lofo_selection}
    \begin{tabular}{lrrrrrr}
        \toprule
        Held-out family &
        Global $U$ &
        Compiler $U$ &
        Oracle $U$ &
        Global Regret &
        Compiler Regret &
        $\Delta R$ \\
        \midrule
        Behavioral
            & .7144 & .7197 & .7534 & .0390 & .0337 & +.0052 \\
        Causal
            & .4252 & .4252 & .5074 & .0822 & .0822 & .0000 \\
        Factual
            & .5044 & .5244 & .5909 & .0865 & .0665 & +.0201 \\
        Lexical
            & .6882 & .6332 & .7898 & .1016 & .1566 & -.0550 \\
        Procedural
            & .4493 & .3635 & .4501 & .0008 & .0866 & -.0858 \\
        \midrule
        Macro
            & .5563 & .5332 & .6183 & .0620 & .0851 & -.0231 \\
        \bottomrule
    \end{tabular}
\end{table}

\section{Experiment 6: Cross-Backbone Replication on Gemma}
\label{app:gemma}

Experiment 6 asks whether the conditions that support adaptation compilation recur on a second model backbone. This appendix provides the complete Gemma protocol, backbone-specific calibration, adaptation geometry, selection headroom, and compiler-selection results.

\subsection{Replication Protocol}
\label{app:gemma_protocol}

We replicate the compiler pipeline on Gemma-2-9B-IT. The learning episodes,
train/validation/test partition, behavioral outcomes, utility function, and
conceptual program library are unchanged from the Llama experiments. The
compiler is retrained from scratch using only Gemma adaptation records; no
Llama predictor parameters or adaptation outcomes are transferred.

The corpus contains 400 meta-training, 100 validation, and 100 held-out test episodes, equally divided across the five learning objectives. Each training and validation episode is adapted once per candidate program using seed 11. Each test episode is adapted under three seeds $\{11, 22, 33\}$, and test geometry is defined by averaging outcomes across
those three adaptations before selection. This yields 3,200 adaptation runs: 1,600 meta-training, 400 validation, and 1,200 test runs.

Gemma contains 42 transformer layers. We instantiate the same conceptual
programs at corresponding relative depths, using layers 0--10 for early
adaptation, 15--25 for middle adaptation, and 31--41 for late adaptation.
All programs target the attention and feed-forward projection modules. 

\subsection{Backbone-Specific Optimization Calibration}
\label{app:gemma_calibration}

Because the study by \citet{ramnauth2026localized} found that Gemma was particularly sensitive to transferring a Llama-calibrated adaptation budget, we calibrate the optimization schedule independently before constructing the Gemma geometry dataset. 

Calibration uses five meta-training episodes from each learning objective
(25 episodes total), seed 11, and a high-capacity full-stack rank-16 adapter. We hold all other optimization choices fixed and vary gradient accumulation over $\{1,2,4,8\}$. No validation or test episodes are used for this decision. 

\begin{table}[t]
\centering
\small
\caption{Gemma optimization-schedule calibration. Values are averaged across the 25 calibration episodes.}
\label{tab:gemma_calibration}
\begin{tabular}{lccccc}
\toprule
Grad. accum. & Acquisition & Transfer & Boundedness & Preservation & Utility \\
\midrule
\rowcolor{black!10} 1 & .560 & .527 & .460 & .998 & .636 \\
2 & .316 & .360 & .400 & .998 & .519 \\
4 & .168 & .207 & .220 & .998 & .398 \\
8 & .068 & .180 & .213 & .998 & .365 \\
\bottomrule
\end{tabular}
\end{table}

No candidate satisfied the pre-specified boundedness gate of .80, so the
automatic calibration rule was formally inconclusive. We do not relax this
criterion post hoc. Instead, after inspecting the calibration tradeoff, we
select gradient accumulation 1 because it dominates the alternatives on acquisition, transfer, and boundedness while preserving essentially identical preservation. This schedule is then fixed for all subsequent Gemma adaptations.

Calibration uses the high-capacity full-stack rank-16 condition to determine whether the optimization schedule supports learning when capacity is not the limiting factor, whereas the primary compiler library uses approximately budget-matched localized rank-16 and full-stack rank-4 programs. Consequently, calibration establishes an executable Gemma-specific schedule rather than guaranteeing that every objective is equally learnable under every primary candidate. We do not recalibrate individual programs after observing their test performance.

\subsection{Gemma Adaptation Geometry}
\label{app:gemma_geometry}

Table~\ref{tab:gemma_geometry} reports the complete test-set geometry after
averaging the three adaptation seeds for each episode--program pair. The
resulting profiles differ substantially across both learning objectives and
adaptation programs. In particular, the mean program profiles show that highest mean utility within a learning family can differ from the program that is oracle-optimal for a substantial fraction of individual episodes. This episode-level heterogeneity is the quantity relevant to compilation.

\begin{table*}[t]
\centering
\small
\caption{Gemma adaptation geometry on held-out test episodes. Values are means
over 20 episodes per objective after averaging the three adaptation seeds.
$A$, $T$, $B$, and $P$ denote acquisition, transfer, boundedness, and
preservation; $U$ is their balanced mean.}
\label{tab:gemma_geometry}
\begin{tabular}{llccccc}
\toprule
Objective & Program & $A$ & $T$ & $B$ & $P$ & $U$ \\
\midrule
Behavioral
 & Early  & .087 & .494 & .000 & .999 & .395 \\
 & Middle & .343 & .556 & .000 & .999 & .474 \\
 & Late   & .092 & .492 & .006 & .971 & .390 \\
 & Full   & .378 & .639 & .014 & .994 & .506 \\
\midrule
Causal
 & Early  & .015 & .036 & .397 & .996 & .361 \\
 & Middle & .057 & .072 & .422 & .997 & .387 \\
 & Late   & .000 & .072 & .356 & .962 & .347 \\
 & Full   & .047 & .111 & .400 & .995 & .388 \\
\midrule
Factual
 & Early  & .027 & .150 & .269 & .995 & .360 \\
 & Middle & .050 & .167 & .478 & .993 & .422 \\
 & Late   & .195 & .369 & .314 & .989 & .467 \\
 & Full   & .142 & .317 & .364 & .992 & .454 \\
\midrule
Lexical
 & Early  & .063 & .147 & .728 & .994 & .483 \\
 & Middle & .138 & .211 & .825 & .998 & .543 \\
 & Late   & .220 & .156 & .783 & .985 & .536 \\
 & Full   & .122 & .175 & .803 & .995 & .524 \\
\midrule
Procedural
 & Early  & .003 & .000 & .000 & .997 & .250 \\
 & Middle & .048 & .000 & .106 & .996 & .287 \\
 & Late   & .015 & .042 & .114 & .955 & .281 \\
 & Full   & .023 & .000 & .219 & .994 & .309 \\
\bottomrule
\end{tabular}
\end{table*}

\subsection{Selection Headroom}
\label{app:gemma_headroom}

As in Experiment 1, we construct fixed baselines using meta-training episodes
only. The global-fixed policy selects full-stack rank 4. Objective-conditioned
defaults select full-stack for behavioral policy, middle for causal mapping,
full-stack for factual association, late for lexical binding, and full-stack
for procedural reasoning.

Table~\ref{tab:gemma_headroom} reports the resulting test performance. Across
all 100 test episodes, the episode-wise oracle achieves mean utility .461,
compared with .436 for the global-fixed policy and .438 for objective-fixed
selection. The corresponding regrets are .025 and .023. Objective conditioning
therefore explains some, but not all, of the variation in preferred adaptation
programs.

\begin{table*}[t]
\centering
\small
\caption{Gemma selection headroom on the 100 held-out test episodes.
Global- and objective-fixed programs are selected using meta-training data
only. ``Opt.'' is the fraction of episodes on which the fixed policy belongs
to the oracle set.}
\label{tab:gemma_headroom}
\resizebox{\linewidth}{!}{
\begin{tabular}{llccccccc}
\toprule
Objective & Obj.-fixed &
Global $U$ & Obj. $U$ & Oracle $U$ &
Global regret & Obj. regret &
Global Opt. & Obj. Opt. \\
\midrule
Overall    & ---    & .436 & .438 & .461 & .025 & .023 & .40 & .53 \\
Behavioral & Full   & .506 & .506 & .522 & .016 & .016 & .75 & .75 \\
Causal     & Middle & .388 & .387 & .399 & .011 & .012 & .30 & .55 \\
Factual    & Full   & .454 & .454 & .492 & .038 & .038 & .35 & .35 \\
Lexical    & Late   & .524 & .536 & .572 & .048 & .036 & .10 & .50 \\
Procedural & Full   & .309 & .309 & .322 & .013 & .013 & .50 & .50 \\
\bottomrule
\end{tabular}
}
\end{table*}

The causal row illustrates why oracle frequency and mean realized utility need
not induce the same ordering: middle adaptation, selected from meta-training
data, is oracle-optimal more frequently than full-stack adaptation on causal
episodes (.55 versus .30), although full-stack attains slightly higher mean
test utility (.388 versus .387). We retain the meta-training-selected middle
default rather than selecting between them using test performance.

\subsection{Oracle-Program Heterogeneity}
\label{app:gemma_oracle_distribution}

Table~\ref{tab:gemma_winners} decomposes the oracle distribution. Using
fractional credit for ties, full, middle, late, and early programs account for 40\%, 31\%, 23\%, and 6\% of episode-level oracle selections, respectively. Only 2\% of test episodes contain an exact top tie. The mean utility difference between the best and second-best program is .039 and the median is .023. Thus, Gemma reproduces the primary prerequisite for compilation which is that there is no single adaptation program, nor even a learning-family default, that eliminates episode-level selection headroom.

\begin{table}[t]
\centering
\small
\caption{Distribution of oracle-optimal Gemma programs on held-out episodes.
Program shares use fractional credit for ties.}
\label{tab:gemma_winners}
\begin{tabular}{lcccc}
\toprule
Objective & Early & Middle & Late & Full \\
\midrule
Overall    & .06 & .31 & .23 & .40 \\
Behavioral & .00 & .25 & .00 & .75 \\
Causal     & .15 & .50 & .05 & .30 \\
Factual    & .00 & .20 & .45 & .35 \\
Lexical    & .10 & .30 & .50 & .10 \\
Procedural & .05 & .30 & .15 & .50 \\
\bottomrule
\end{tabular}
\end{table}

\subsection{Compiler Selection on Gemma}
\label{app:gemma_compiler}

The Gemma geometry predictor is trained only on the 400 Gemma meta-training
episodes, with model class and hyperparameters selected on the 100 Gemma
validation episodes. The predictor receives the same full pre-adaptation
episode--model representation used in the primary Llama experiment and does
not receive learning-objective identity.

%

Table~\ref{tab:gemma_compiler_selection} reports downstream selection. The
compiler achieves mean utility .435 versus .461 for the episode-wise oracle,
corresponding to .026 mean regret. It recovers an oracle-optimal program on
42\% of episodes and an oracle top-two program on 73\%.

\begin{table*}[t]
\centering
\small
\caption{Gemma compiler selection by learning objective. ``Selections''
reports the number of the 20 held-out episodes assigned to each program.
Top-1 and Top-2 are oracle recovery rates.}
\label{tab:gemma_compiler_selection}
\resizebox{\linewidth}{!}{
\begin{tabular}{lllcccccc}
\toprule
Objective & Obj.-fixed & Compiler selections &
Compiler $U$ & Obj. $U$ & Oracle $U$ &
Compiler regret & Top-1 & Top-2 \\
\midrule
Behavioral & Full   & Full: 20
    & .506 & .506 & .522 & .016 & .75 & 1.00 \\
Causal & Middle     & Full: 20
    & .388 & .387 & .399 & .011 & .30 & .55 \\
Factual & Full      & Full: 20
    & .454 & .454 & .492 & .038 & .35 & .85 \\
Lexical & Late      & Full: 10; Late: 10
    & .532 & .536 & .572 & .040 & .35 & .55 \\
Procedural & Full   & Full: 8; Middle: 12
    & .297 & .309 & .322 & .025 & .35 & .70 \\
\midrule
Overall & --- & ---
    & .435 & .438 & .461 & .026 & .42 & .73 \\
\bottomrule
\end{tabular}
}
\end{table*}

Relative to objective-fixed selection, the compiler improves realized utility
on 14 episodes, decreases it on 28, and makes an equivalent choice on the
remaining 58. Its mean utility is therefore slightly below the objective-fixed
policy (.435 versus .438) and essentially matches the global-fixed policy
(.436).

\subsection{Where Does Gemma Selection Fail?}
\label{app:gemma_failure_analysis}

The aggregate difference is highly concentrated rather than uniform across
learning objectives. The compiler exactly reproduces the full-stack default on
all behavioral and factual episodes. On causal mapping it also always selects
full-stack; this yields slightly higher mean test utility than the
meta-training-selected middle default, although lower oracle recovery.
The net shortfall relative to the objective-fixed policy therefore comes from
lexical and, especially, procedural episodes.

For lexical binding, the compiler divides its selections evenly between full
and late adaptation. These switches improve utility relative to the global
full-stack policy but underperform the objective-level late default on average.
For procedural reasoning, the compiler selects middle adaptation on 12 of 20
episodes even though the meta-training default is full-stack. This reduces mean
utility from .309 under the fixed full-stack policy to .297. Procedural
reasoning alone accounts for approximately 79\% of the compiler's aggregate
utility deficit relative to objective-fixed selection; the smaller lexical
deficit is partly offset by a gain on causal episodes.

The procedural errors also reveal a decision-calibration issue. Among the
seven procedural episodes for which the compiler selects middle while
full-stack is the observed oracle, the predicted advantage of middle over the
oracle full-stack program is only .0027 on average (median .0031), yet the
mean realized regret of making that switch is .050. Thus, the argmax decision
rule can act on predicted differences that are small relative to the
consequence of choosing the wrong program.

\begin{table}[t]
\centering
\small
\caption{Diagnostic for procedural episodes on which the compiler selects
middle adaptation while full-stack is oracle-optimal. The predicted margin is
the predicted utility advantage that triggers the middle-program selection.}
\label{tab:gemma_margin_diagnostic}
\begin{tabular}{lc}
\toprule
Statistic & Value \\
\midrule
Number of episodes & 7 \\
Mean predicted margin & .0027 \\
Median predicted margin & .0031 \\
Mean realized regret & .0501 \\
\bottomrule
\end{tabular}
\end{table}

This distinction clarifies the Gemma result. Episode-specific adaptation
headroom remains present, but exploiting that headroom requires not only
accurate expected geometry but sufficiently reliable discrimination between
nearby candidate programs. The current compiler treats any predicted utility
advantage as actionable, regardless of its magnitude.

\subsection{Relationship to the Llama Results}
\label{app:gemma_llama_comparison}

The Gemma replication does not support an architecture-invariant program map.
Table~\ref{tab:backbone_defaults} compares the objective-conditioned
meta-training defaults across backbones. Only causal mapping selects the same
region in both models.

\begin{table}[t]
\centering
\small
\caption{Objective-conditioned adaptation defaults across backbones. Defaults
are estimated independently from each backbone's meta-training episodes.}
\label{tab:backbone_defaults}
\begin{tabular}{lcc}
\toprule
Objective & Llama & Gemma \\
\midrule
Behavioral policy    & Late   & Full \\
Causal mapping       & Middle & Middle \\
Factual association  & Late   & Full \\
Lexical binding      & Early  & Late \\
Procedural reasoning & Middle & Full \\
\bottomrule
\end{tabular}
\end{table}

This result is consistent with the earlier cross-model finding that adaptation
geometry contains both objective-specific and model-specific structure
\citep{ramnauth2026localized}. More importantly for the present paper, the
\emph{selection problem} itself reproduces: multiple programs remain
oracle-optimal across individual Gemma episodes and objective-specific
defaults leave measurable residual regret.

What does not reproduce unchanged is the sufficiency of a deterministic
argmax over predicted geometry. On Llama, episode-level predictions are
accurate enough that direct compilation improves over both fixed policies. On
Gemma, the same selection rule sometimes responds to small predicted
differences that do not justify departing from a strong default. The
cross-backbone result therefore suggests that successful adaptation
compilation has two requirements: (1) candidate programs must expose meaningful episode-specific headroom, and (2) the compiler must be sufficiently calibrated to know when that headroom can be exploited reliably.

A natural extension is therefore confidence-aware compilation, in which a
compiler retains a robust fixed default unless the predicted advantage of an alternative program exceeds a threshold or uncertainty criterion estimated from validation data. We do not introduce such a rule post hoc here; the Gemma results are reported using the same direct selection procedure specified for the primary experiments.

\end{document}